\documentclass[11pt,twoside]{article}

\usepackage{palatino, geometry, url}

\usepackage{cite}
\usepackage{amsmath,amssymb,amsfonts}
\usepackage{graphicx}
\usepackage{array}
\usepackage{tabularx}
\usepackage{makecell}
\usepackage{multirow}

\allowdisplaybreaks

\usepackage{fancyhdr}

\usepackage{cite}
\usepackage{graphicx}
\usepackage{textcomp}
\usepackage{wrapfig}
\usepackage{amsmath,amsfonts}
\usepackage{algorithmic}
\usepackage{algorithm}
\usepackage{array}
\usepackage{booktabs}    
\usepackage{textcomp}
\usepackage{stfloats}
\usepackage{url}
\usepackage{verbatim}
\usepackage{graphicx}
\usepackage{caption}
\usepackage{xcolor}
\usepackage{float}

\usepackage[justification=justified,singlelinecheck=false]{caption}
\usepackage{subcaption}

\makeatletter
\let\NAT@parse\undefined
\makeatother

\usepackage[hidelinks]{hyperref}

\begin{document}

\title{Physics-Guided Conditional Diffusion Model for Rare Event Synthesis and Diagnosis for the Water-Gas Shift Reaction}

\author{Md Abrar Rafid Siddique${}^\dag$, Bibek Aryal${}^\dag$, and Qiugang Lu${}^{\dag}$
\thanks{Corresponding author: Qiugang Lu; Email: jay.lu@ttu.edu.}\\
\\
{\small ${}^\dag$Department of Chemical Engineering, Texas Tech University, Lubbock, TX 79409, USA}
\\
}

\date{}
\maketitle

\begin{abstract}
As the world moves towards sustainable energy sources, hydrogen (H\textsubscript{2}) can be treated as an eco-friendly alternative to fossil fuels due to its high energy density and zero carbon emission. The water-gas shift (WGS) reaction is a widely used industrial process for hydrogen production by converting carbon monoxide and steam into  hydrogen and carbon dioxide. However, occurrences like severe fouling, catalyst deterioration, and thermal runaway can hamper the reaction kinetics/process safety and decrease the yield of H\textsubscript{2}. These incidents are rare, and gathering process data under such abnormal conditions is challenging. In this work, we propose a physics-guided conditional diffusion model to generate realistic rare-event trajectories for the WGS reaction. The proposed model integrates a conditional denoising diffusion probabilistic model (CDDPM) with governing laws of the reaction to generate physically consistent process trajectories. The conditioning features allow the model to produce high-quality synthetic profiles for rare-event domains that are typically beyond the training regimes. The generated rare-event trajectories then augment the raw dataset for a balanced distribution between normal and abnormal conditions. We further propose a hazard score to assess the risk severity of the operating condition based on the operating trajectory. Deep learning models are trained with the augmented dataset to diagnose the health status of the reaction. Simulation results show that the proposed physics-guided diffusion model outperforms data-driven models in terms of the quality of synthetic data and diagnosis performance for rare events.

\end{abstract}

\noindent\textbf{Keywords:} Water-gas shift reaction; Physics-guided conditional diffusion model; Rare-event data synthesis; Hazard score index; Process fault diagnosis.

\section{Introduction}

Excessive use of fossil fuels has increased global warming in recent years \cite{bilgili2024comprehensive,hosseini2022fossil}. During combustion, fossil fuels emit large amounts of greenhouse gases into the atmosphere. Greenhouse gases (CH\textsubscript{4}, CO\textsubscript{2}, N\textsubscript{2}O) absorb infrared radiation and re-emit it into the earth's atmosphere, causing global warming \cite{rezaei2025global,wang2024natural}. Unlike fossil fuels, hydrogen (H\textsubscript{2}) produces only water vapour (H\textsubscript{2}O) during combustion and does not emit greenhouse gases. Therefore, hydrogen can be considered as a clean and sustainable energy source alternative to fossil fuels \cite{patel2025catalysts,midilli2005hydrogen}. Among various hydrogen production technologies, the water-gas shift (WGS) reaction is widely used as a benchmark process. In this reaction, carbon monoxide (CO) reacts with water vapour (H\textsubscript{2}O) to produce hydrogen (H\textsubscript{2}) and carbon dioxide (CO\textsubscript{2}) with the aid of catalysts. As a result, the hydrogen content in syngas (a mixture of CO and H\textsubscript{2}) increases while the carbon monoxide content decreases \cite{patel2025catalysts,dehimi2025hydrogen,lee2023advances}.

WGS reactions are normally operated within safe limits. However, under severe operating conditions, catalyst deactivation may occur due to thermal sintering, sulphur poisoning, and chloride poisoning \cite{baraj2021water,beale2014chemical}. In addition, severe fouling over time can reduce heat transfer efficiency, increasing the risk of thermal runaway because the WGS reaction is moderately exothermic \cite{bac2018modeling}. Since these abnormal conditions occur rarely, only highly limited operating data under these conditions, if any, are available. This creates significant challenges for reliable monitoring, diagnosis, and mitigation of abnormal operations for WGS reactions to prevent disasters from occurring. 

To overcome this issue, various data augmentation methods have been developed by creating high-fidelity synthetic profiles under rare events. A common strategy is the numerical simulation of reaction models by leveraging explicit governing laws, such as mass, energy balance, and reaction kinetics to generate numerical data for rare scenarios. For instance, Chen et al.\cite {chen2008modeling} simulated the WGS reaction via discretization of such governing laws with finite-volume method followed by semi-explicit algorithms. Bac et al. \cite {bac2018modeling} modeled the WGS process with 2D Navier-Stokes equations alongside reactive heat and mass transport. They incorporated  heat exchange functions to minimize the impact of thermodynamic limitations to increase the overall conversion of CO. However, these methods require full knowledge of the underlying governing laws that may not be available for many processes and operating conditions.

On the other hand, generative models, such as generative adversarial networks (GANs) and diffusion models, have emerged for data synthesis to overcome the scarcity issue \cite{goodfellow2014generative, ho2020denoising}. Specifically, GANs involve an adversarial training between a discriminator and a generator, and thus often suffer from  training instability and mode collapse \cite{zheng2026filtered,li2025population}. Unlike GANs, diffusion models start with random noise and iteratively denoise it until it forms a realistic sample that resembles the data that they were trained on \cite{ho2020denoising,nichol2021improved}. The absence of adversarial training allows the diffusion model to avoid issues like mode collapse \cite{zheng2026filtered}. Thus, diffusion models have been widely adopted in the synthesis of various data forms such as images, audio, and time series  \cite{rombach2022high,kong2020diffwave,zheng2026filtered,shen2023non,li2025population}. For chemical processes, where operating data are predominantly time series, diffusion models have been reported to generate synthetic data for reactions such as the ozone–nitric oxide reaction \cite{millard2026particle}, the prediction of gas-dispersion field distributions \cite{chen2026spatiotemporal}, and material design and drug discovery \cite{alakhdar2024diffusion,wang2025diffusion,guo2024diffusion}.


Although diffusion models are effective for time-series data synthesis, they often overlook the system governing laws from which time-series data are collected. Thus, how to enable the physical plausibility of synthetic data is of priority for time-series data synthesis. To this end, various physics-informed diffusion models have been proposed. For instance, Yuan et al. \cite{yuan2023physdiff} proposed a physics-guided motion diffusion model, PhysDiff, and incorporated physical constraints to iteratively pull the motion toward a physically plausible space. Wang et al. \cite{wang2025phyda} proposed PhyDA, a physics-guided diffusion model for data assimilation in atmospheric systems. They incorporated a partial differential equation (PDE)-based physics loss during the training to generate physically consistent atmospheric data. For chemical processes such as WGS-based H$_2$ production, the compliance with physical laws is critical, since otherwise the synthetic data may show unrealistic behaviors such as negative flow rates, unreasonable temperatures, among others. Moreover, the inclusion of physics can assist the generative model to overcome data scarcity and extrapolate to regimes beyond the training data domain \cite{yang2026least}. This is crucial for data synthesis of rare-event cases to enable effective monitoring and prevention of such scenarios. However, to our best knowledge, there has been no reports on developing physics-guided diffusion models for rare-event data synthesis and diagnosis for the WGS reaction, which motivates this work. 

\begin{figure}[!ht]
	\centering
	\makebox[\textwidth][c]{%
		\includegraphics[width=0.9\textwidth]{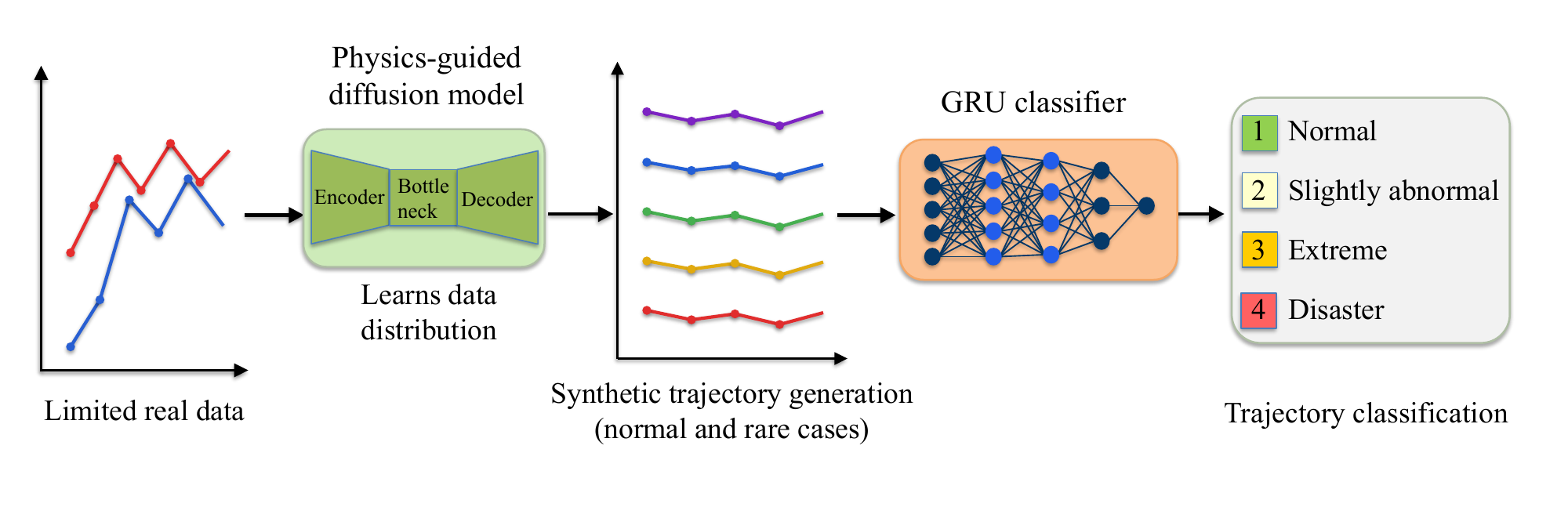}
	}
	\caption{The overall framework of the proposed Pg-CDDPM-based rare-event synthesis and diagnosis method.}
	\label{fig:framework}
\end{figure}

In this study, we propose a physics-guided conditional denoising diffusion probabilistic model (Pg-CDDPM) for rare-event data synthesis and fault diagnosis of the WGS process. The proposed model incorporates physics (may not be exact) loss into the conventional noise prediction loss to ensure that the generated trajectories satisfy the governing equations. This physics-informed nature together with the conditioning feature allow for the extrapolation to generate rare-event synthetic data. The overall framework of our method is shown in Fig. ~\ref{fig:framework}, where limited trajectory data are used by Pg-CDDPM to generate additional synthetic trajectories for rare-event conditions, followed by training a monitoring model (gated recurrent unit, GRU) to diagnose the severity level of abnormal operation in the test data. The main contributions are summarized as follows:
\begin{itemize}
\item The WGS reaction and reactor dynamics are modeled using the Eley–Rideal reaction mechanism and a continuous stirred-tank reactor (CSTR) model. Critical parameters, e.g., heat transfer coefficient and activation energy factor, are selected and adjusted to create different operating conditions from normal to abnormal and ultimately disaster scenarios. 
	
\item A Pg-CDDPM with a unique backbone 1D U-Net architecture is proposed that incorporates governing equations and is conditioned on heat transfer efficiency and activation energy factors to generate physically consistent WGS operating profiles under normal and rare-event operating conditions, including both interpolation and extrapolation regimes.
	
	\item A composite health indicator is developed to  assess the health status of the reaction. Different levels of severity are defined based on the health indicator, including normal, slightly abnormal, extreme, and disaster. The generated synthetic data for rare-event scenarios (extreme and disaster) are leveraged to train the health monitoring model for future diagnosis. 
		

	\item Extensive validations based on the WGS simulator are conducted to evaluate the fidelity of synthetic data, as well as the effectiveness of augmenting raw data with synthetic data in improving the diagnosis performance for different severity classes. 
	
\end{itemize}


\section{Preliminaries}
\label{sec: Preliminaries}
\subsection{Water-Gas Shift Reaction }
The WGS reaction is widely used in the industry for hydrogen production and ammonia synthesis. In this reversible reaction,
carbon monoxide (CO) reacts with water vapor (H$_2$O) to produce hydrogen
(H$_2$) and carbon dioxide (CO$_2$):
\begin{equation}
	\mathrm{CO + H_2O \rightleftharpoons CO_2 + H_2},
	\qquad
	\Delta H = -41.1~\mathrm{kJ/mol}.
	\label{eq:wgs}
\end{equation}
The reaction is moderately exothermic, and its equilibrium conversion decreases with increasing temperature. The reaction is favored thermodynamically at lower temperatures and kinetically at elevated temperatures. Industrially, the reaction is conducted in two stages: a high temperature stage and a low temperature stage. In a high temperature shift reaction, the reactor operates between 300$^\circ$C and 550$^\circ$C and uses an iron-based catalyst (Fe/Cr). In contrast, in a low temperature stage, the reactor operates between 150$^\circ$C and 230$^\circ$C and uses a copper–zinc catalyst supported over alumina (CuO/ZnO/Al$_2$O$_3$) \cite{degliuomini2012rigorous,baraj2021water,patel2025catalysts}. As the equilibrium CO conversion decreases at very high temperatures, an intermediate cooling system is included to maintain the temperature within the desired operating range to increase the overall CO conversion.

\subsection{Denoising Diffusion Probabilistic Models}
\begin{figure}[!ht]
	\centering
	\makebox[\textwidth][c]{%
		\includegraphics[width=0.7 \textwidth]{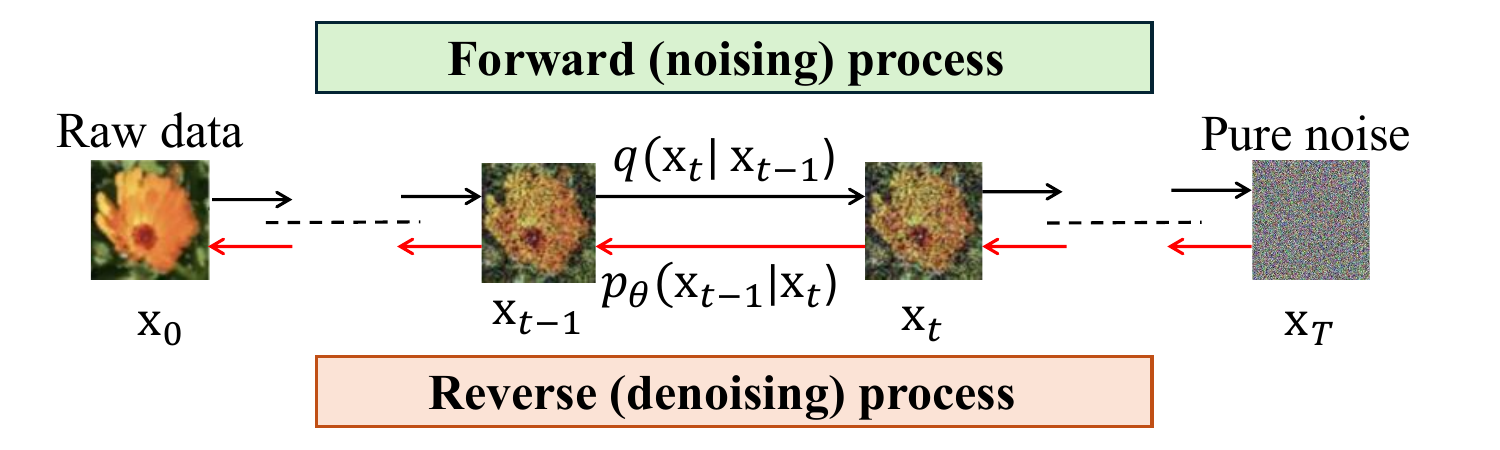}
	}
	\caption{Overview of the forward and reverse processes of a diffusion model.}
	\label{fig:diffusion}
\end{figure}

Diffusion models are a class of probabilistic generative models. They learn the underlying data distribution through a two-stage process. First, they destroy the original data by progressively adding noises. In the reverse process, they learn through a denoising process to generate new samples from pure noise \cite{ho2020denoising,nichol2021improved,zheng2026filtered}. Among various diffusion models, the DDPM proposed by Ho et al. \cite{ho2020denoising} is a common type due to its ability to produce high-quality samples. As in Fig.~\ref{fig:diffusion}, the DDPM consists of two Markov chains: a forward diffusion process and a reverse diffusion process. In the forward process, Gaussian noise is gradually added to the original data until it approaches pure noise. In the reverse process, a neural network learns to estimate the added noise and progressively removes it, allowing new data samples to be generated from pure noise. Denote the original data sample with true distribution as $\mathrm{x}_0 \sim q(\mathrm{x}_0)$. The diffusion process consists of $T$ time steps, with each diffusion step $t \in \{1,2,\ldots,T\}$. The forward process gradually adds Gaussian noise to the original data according to a predefined variance schedule $\{\beta_t\}_{t=1}^{T}$, expressed as \cite{ho2020denoising}
\begin{equation}
	q(\mathrm{x}_{1:T}\mid \mathrm{x}_0)
	=
	\prod_{t=1}^{T}
	q(\mathrm{x}_t\mid\mathrm{x}_{t-1}),
	\label{eq:forward}
\end{equation}
\begin{equation}
	q(\mathrm{x}_t\mid\mathrm{x}_{t-1})
	=
	\mathcal{N}
	\left(
	\mathrm{x}_t;
	\sqrt{1-\beta_t}\,\mathrm{x}_{t-1},
	\beta_t\mathrm{I}
	\right),
	\label{eq:forward_transition}
\end{equation}
where $\mathrm{x}_t$ is the noisy sample at diffusion step $t$, $\beta_t$ is a predefined variance schedule typically increasing over $t$, and $\mathrm{I}$ is the identity matrix. Define
\begin{equation}
	\alpha_t = 1-\beta_t,
	\qquad
	\bar{\alpha}_t=\prod_{i=1}^{t}\alpha_i.
	\label{eq:alpha}
\end{equation}
Then the distribution of $\mathrm{x}_t$ conditioned directly on the original sample $\mathrm{x}_0$ can be written as
\begin{equation}
q(\mathrm{x}_t\mid\mathrm{x}_0)
=
\mathcal{N}
\left(
\mathrm{x}_t;
\sqrt{\bar{\alpha}_t}\,\mathrm{x}_0,
(1-\bar{\alpha}_t)\mathrm{I}
\right).
\label{eq:forward_closed}
\end{equation}
Consequently, the noisy sample at any diffusion time step $t$ becomes
\begin{equation}
	\mathrm{x}_t
	=
	\sqrt{\bar{\alpha}_t}\,\mathrm{x}_0
	+
	\sqrt{1-\bar{\alpha}_t}\,\epsilon,
	\qquad
	\epsilon\sim\mathcal{N}(0,\mathrm{I}),
	\label{eq:xt}
\end{equation}
where ${\epsilon}$ is the standard Gaussian noise. This formulation enables the noisy sample at any arbitrary $t$ to be generated directly from the original data without repeatedly applying the forward diffusion process. The reverse process starts from a pure Gaussian noise sample $\mathrm{x}_T \sim \mathcal{N}(\mathrm{0},\mathrm{I})$. The objective is to gradually remove the added noise and construct new samples that follow the original data distribution. It is defined as a learnable Markov chain via a parameterized neural network
\begin{equation}
	p_{\theta}(\mathrm{x}_{0:T})
	=
	p(\mathrm{x}_T)
	\prod_{t=1}^{T}
	p_{\theta}(\mathrm{x}_{t-1}\mid\mathrm{x}_t),
	\label{eq:reverse_joint}
\end{equation}
where $p(\mathrm{x}_T)	=
	\mathcal{N}(\mathrm{x}_T;\mathrm{0},\mathrm{I}). $
Each reverse transition is modeled as
\begin{equation}
	p_{\theta}
	\left(
	\mathrm{x}_{t-1}\mid\mathrm{x}_t
	\right)
	=
	\mathcal{N}
	\left(
	\mathrm{x}_{t-1};
	\mu_{\theta}(\mathrm{x}_t,t),
	\Sigma_{\theta}(\mathrm{x}_t,t)
	\right).
	\label{eq:reverse_transition}
\end{equation}
The network learns to predict the noise $\epsilon_{\theta}(x_t,t)$ added during the forward process, and then
\begin{equation}
	\mathrm{x}_{t-1}
	=
	\frac{1}{\sqrt{\alpha_t}}
	\left(
	\mathrm{x}_t
	-
	\frac{\beta_t}{\sqrt{1-\bar{\alpha}_t}}
	\epsilon_{\theta}(\mathrm{x}_t,t)
	\right)
	+
	\sqrt{\beta_t}\,z,
	\qquad
	z\sim\mathcal{N}(0,I),
	\label{eq:reverse_sample}
\end{equation}
where $z$ is standard Gaussian noise. Starting from a pure noise sample $\mathrm{x}_T$, Eq.~(9) is repeatedly applied from $t=T$ to $t=1$ to progressively remove the added noise and generate a sample that follows the original data distribution \cite{ho2020denoising}.

\section{Methodology}
\label{sec: Methodology}
\subsection{WGS Simulator}
\begin{figure}[!ht]
	\centering
	\makebox[\textwidth][c]{%
		\includegraphics[width=1\textwidth]{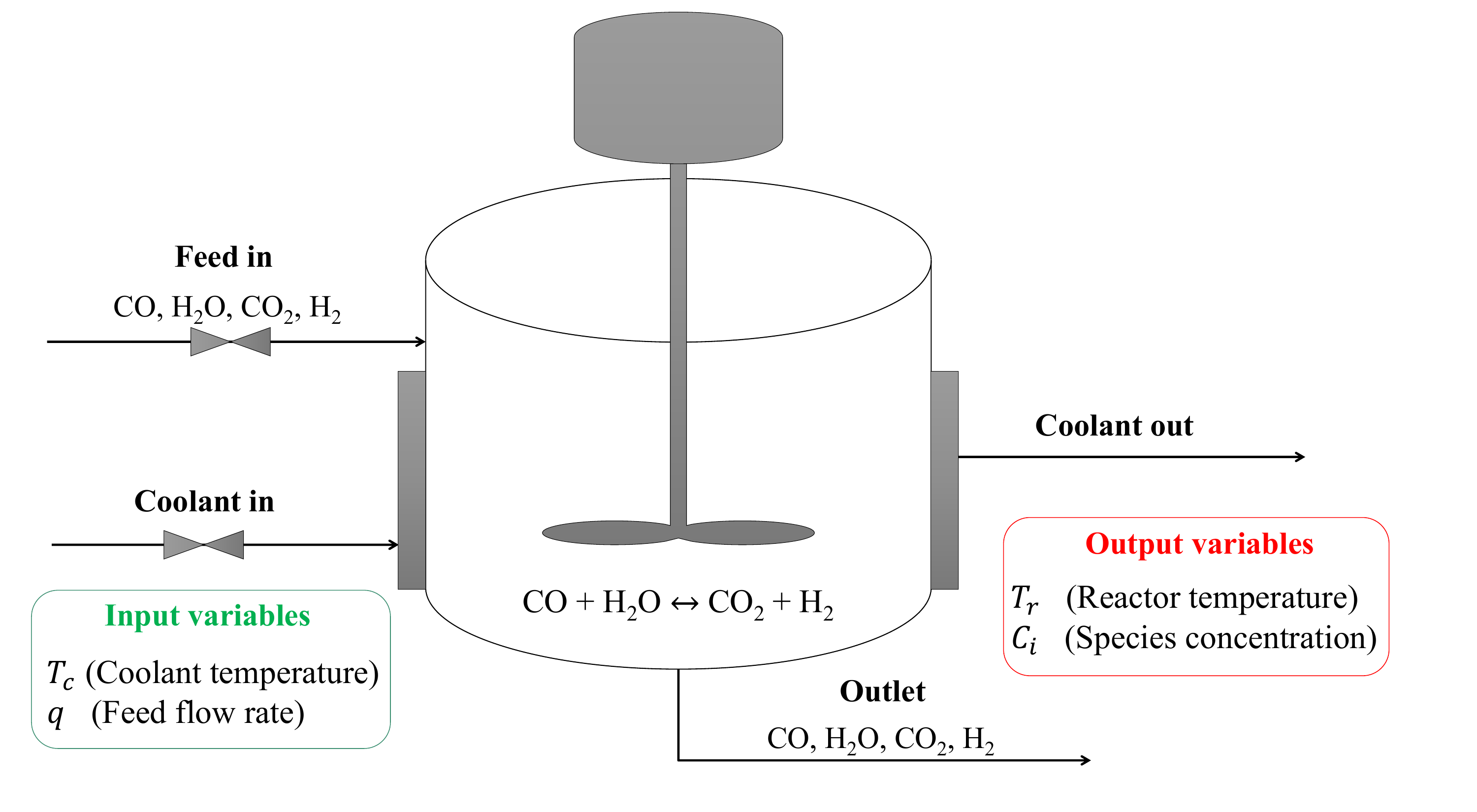}
	}
	\caption{Schematic diagram of the WGS reaction in a continuous stirred-tank reactor.}
	\label{fig:cstr}
\end{figure}

A dynamic simulator of the WGS reaction Eq. \eqref{eq:wgs} is developed to generate operating trajectories of process variables for training and evaluating the proposed Pg-CDDPM. The simulator captures transient behaviors of the reaction by solving governing equations under varying heat transfer efficiency and activation energy factors. A schematic diagram of the CSTR is shown in Fig.~\ref{fig:cstr}. We assume that the reactor has perfect mixing, constant volume, and negligible pressure drop. Specifically, the reaction rate is described by the Eley--Rideal kinetic model with temperature-dependent reaction rate constants obtained from the Arrhenius equation. Reaction dynamics are governed by species mole balances for CO, H$_2$O, CO$_2$, and H$_2$, alongside the energy balance that accounts for the heat released by the reaction and heat exchange with the cooling system. In particular, the mole balance equations are:
\begin{align}
	\frac{dC_{\mathrm{CO}}}{dt}
	&=
	\frac{F}{V}\left(C_{\mathrm{CO},f}-C_{\mathrm{CO}}\right)-r,
	\\
	\frac{dC_{\mathrm{H_2O}}}{dt}
	&=
	\frac{F}{V}\left(C_{\mathrm{H_2O},f}-C_{\mathrm{H_2O}}\right)-r,
	\\
	\frac{dC_{\mathrm{CO_2}}}{dt}
	&=
	\frac{F}{V}\left(C_{\mathrm{CO_2},f}-C_{\mathrm{CO_2}}\right)+r,
	\\
	\frac{dC_{\mathrm{H_2}}}{dt}
	&=
	\frac{F}{V}\left(C_{\mathrm{H_2},f}-C_{\mathrm{H_2}}\right)+r,
\end{align}
where $C_i$ and $C_{f,i}$ denote the reactor and feed concentrations of species $i$, respectively, $F$ is the volumetric flow rate, $V$ is the reactor volume, and $r$ is the reaction rate (to be defined in Section 3.2). With a cooling water system, the energy balance equation can be expressed as:
\begin{equation}
	\frac{dT_r}{dt}
	=
	\frac{F}{V}\left(T_f-T_r\right)
	-\frac{\Delta H_r}{\rho C_p}r
	+\frac{UA}{V\rho C_p}\left(T_c-T_r\right),
	\label{eq:energy_balance}
\end{equation}
where $T_r$ and $T_f$ are the reactor and feed temperatures, respectively, $T_c$ is the coolant temperature, $\Delta H_r$ is the heat of reaction, $\rho$ is the fluid density, $C_p$ is the heat capacity, and $UA$ is the overall heat transfer coefficient. For this reaction, the input variables include the coolant temperature $T_c$ and feed flow rate $F$, whereas the output variables include reactor temperature $T_r$ and concentration of species $C_i$.

\subsection{Reaction Mechanism}
An iron-based catalyst is selected to develop the reaction rate expression. The catalyst operates in a temperature range of $300$--$530^{\circ}\mathrm{C}$. Based on the literature, the WGS reaction on iron-based catalysts follows the Rideal--Eley mechanism, in which steam reacts with an active catalytic site to form hydrogen and an adsorbed oxygen species. The adsorbed oxygen then reacts with carbon monoxide to produce carbon dioxide while regenerating the active sites. The elementary reaction steps are given by \cite{Davis2003}
\begin{equation}
	\mathrm{H_2O + *} \rightleftharpoons \mathrm{H_2 + O^*}, ~\quad
	\mathrm{CO + O^*} \rightleftharpoons \mathrm{CO_2 + *}, \nonumber
\end{equation}
where $*$ denotes an active catalytic site and $O^*$ represents an oxygen species adsorbed on the catalyst surface. The Eley--Rideal rate expression is derived based on the following assumptions: (i) the adsorption step is at quasi-equilibrium (steady state); (ii) the surface reaction is the rate-determining step; and
(iii) the catalyst active sites are regenerated after product desorption. Accordingly, the reaction-rate expression adopted for the reactor model is
\begin{equation}
	r =
	\frac{k_2 K_1 C_{\mathrm{CO}} C_{\mathrm{H_2O}}
		-
		k_{2,\mathrm{rev}} C_{\mathrm{CO_2}} C_{\mathrm{H_2}}}
	{C_{\mathrm{H_2}} + K_1 C_{\mathrm{H_2O}}},~~~	k_2 = k_0 \exp\!\left(-\frac{E_a}{RT_r}\right).
	\label{eq:reaction_rate}
\end{equation}
For the forward reaction rate constant $k_2$, $k_0$ is the pre-exponential factor, $E_a$ is the activation energy, $R$ is the gas constant, and $T_r$ is the reactor temperature.  The reverse reaction rate constant is assumed to be proportional to the forward rate constant, $k_{2,\mathrm{rev}} = 0.5\,k_2$, with the adsorption equilibrium constant taken as $K_1 = 0.5$. 

In our study, we use effective heat transfer coefficient $UA_{eff}$ and effective activation energy $E_{a,eff}$ to replace those in \eqref{eq:energy_balance} and \eqref{eq:reaction_rate}, respectively, so as to better represent the actual heat transfer efficiency and catalyst activity observed during practical operations:
\begin{equation}
UA_{eff}=f\cdot UA,\quad E_{a,eff}=f_{E_a}\cdot E_a, \label{eq: arrhenius}
\end{equation}
where $f$ and $f_{E_a}$ stand for heat transfer efficiency factor and activation energy factor. These two factors will be are carefully varied to create different operating conditions and generate diverse process variable trajectories.



\begin{figure}[!ht]
	\centering
	\makebox[\textwidth][c]{%
		\includegraphics[width=0.9\textwidth]{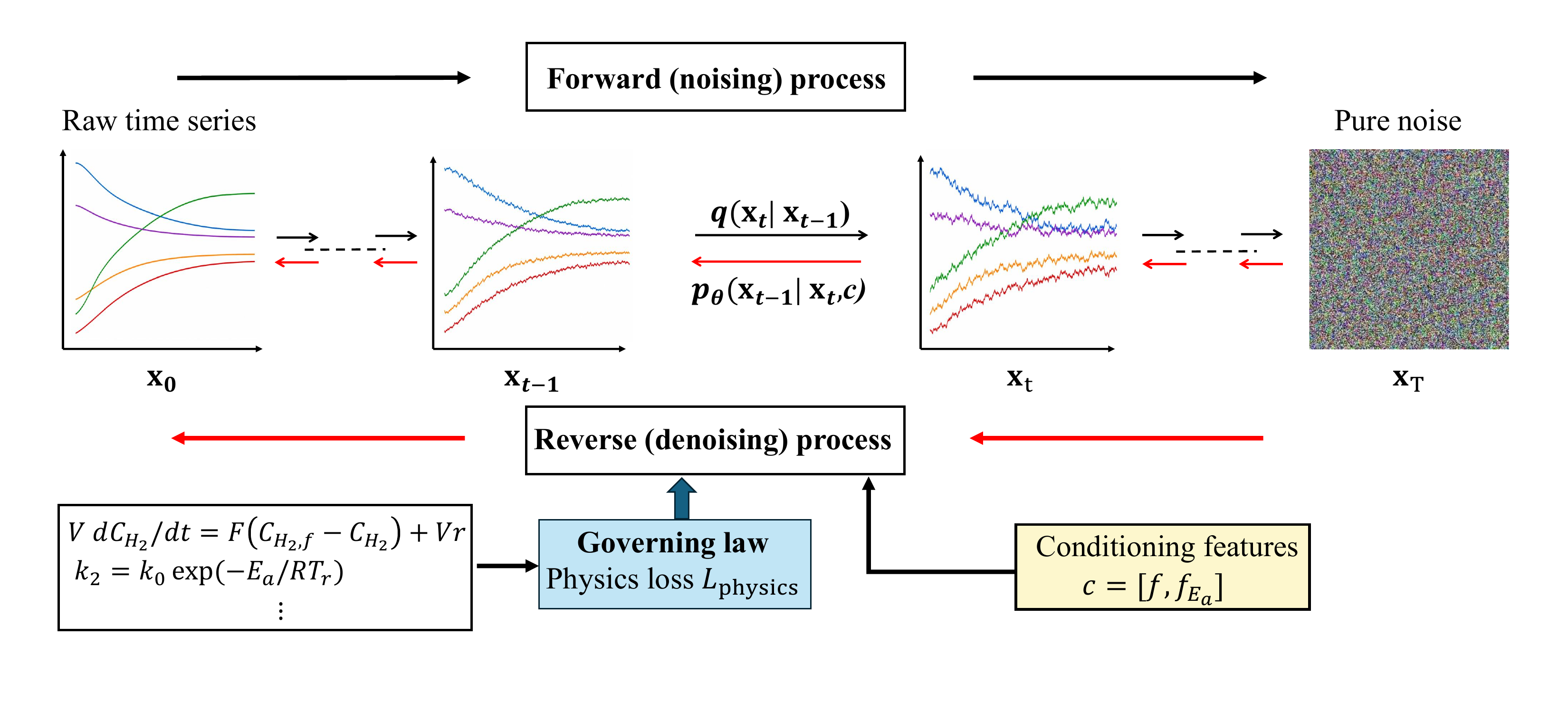}
	}
	\caption{Overview of the proposed Pg-CDDPM framework.}
	\label{fig:diffusion_time_series}
\end{figure}

\subsection{Proposed Physics-Guided Conditional DDPM}
In our study, we will develop a novel Pg-CDDPM to generate synthetic data for the above WGS reaction, illustrated in Fig.~\ref{fig:diffusion_time_series}. First, the original multivariate trajectory $\mathrm{x}_{1:L}^{0}$ is transformed into pure noise $\mathrm{x}_{1:L}^{T}$ through the forward diffusion process, with its reconstruction through the reverse denoising process. During the reverse process, the governing equations are incorporated as physics guidance to improve the physical consistency of generated trajectories. As in Section 2.2, the forward process is defined as
\begin{equation}
	q\!\left(\mathrm{x}_{1:L}^{1:T}\mid\mathrm{x}_{1:L}^{0}\right)
	=
	\prod_{t=1}^{T}
	q\!\left(
	\mathrm{x}_{1:L}^{t}
	\mid
	\mathrm{x}_{1:L}^{t-1}
	\right),~~q\!\left(
	\mathrm{x}_{1:L}^{t}
	\mid
	\mathrm{x}_{1:L}^{t-1}
	\right)
	=
	\mathcal{N}
	\left(
	\mathrm{x}_{1:L}^{t};
	\sqrt{1-\beta_t}\,\mathrm{x}_{1:L}^{t-1},
	\beta_t\mathrm{I}
	\right),
	\label{eq:forward_process}
\end{equation}
where \(T\) is the total number of diffusion time steps, \(L\) is the total trajectory length, \(\mathrm{x}_{1:L}^{0}\) is the clean (ground-truth) raw trajectory, and \(\mathrm{x}_{1:L}^{t}\) is the corresponding noisy trajectory at diffusion step \(t\). With Eq. \eqref{eq:alpha}, the noisy trajectory at diffusion step $t$ can be sampled directly from the original clean trajectory without iteratively applying the forward process:
\begin{equation}
	q\!\left(
	\mathrm{x}_{1:L}^{t}
	\mid
	\mathrm{x}_{1:L}^{0}
	\right)
	=
	\mathcal{N}
	\left(
	\mathrm{x}_{1:L}^{t};
	\sqrt{\bar{\alpha}_t}\,\mathrm{x}_{1:L}^{0},
	\left(1-\bar{\alpha}_t\right)\mathrm{I}
	\right).
	\label{eq:forward_closed}
\end{equation}
The noisy trajectory at any diffusion time step $t$ can be computed as
\begin{equation}
	\mathrm{x}_{1:L}^{t}
	=
	\sqrt{\bar{\alpha}_t}\,\mathrm{x}_{1:L}^{0}
	+
	\sqrt{1-\bar{\alpha}_t}\,\epsilon_{1:L},
	\qquad
	\epsilon_{1:L}\sim\mathcal{N}(0,I).
	\label{eq:forward_sample}
\end{equation}
For \textit{conditional} diffusion models, the reverse process progressively removes the added noise for conditioning feature $c$:
\begin{equation}
	p_{\theta}\!\left(
	\mathrm{x}_{1:L}^{0:T}\mid\mathrm{c}
	\right)
	=
	p\!\left(\mathrm{x}_{1:L}^{T}\right)
	\prod_{t=1}^{T}
	p_{\theta}\!\left(
	\mathrm{x}_{1:L}^{t-1}
	\mid
	\mathrm{x}_{1:L}^{t},\mathrm{c}
	\right).
	\label{eq:reverse_process}
\end{equation}
where 
\begin{equation}
	p_{\theta}\!\left(
	\mathrm{x}_{1:L}^{t-1}
	\mid
	\mathrm{x}_{1:L}^{t},\mathrm{c}
	\right)
	=
	\mathcal{N}\!\left(
	\mathrm{x}_{1:L}^{t-1};
	\mu_{\theta,1:L}\!\left(
	\mathrm{x}_{1:L}^{t},t,\mathrm{c}
	\right),
	\Sigma_{\theta,1:L}\!\left(
	\mathrm{x}_{1:L}^{t},t,\mathrm{c}
	\right)
	\right).
	\label{eq:reverse_transition}
\end{equation}
\begin{equation}
	\mathrm{x}_{1:L}^{t-1}
	=
	\frac{1}{\sqrt{\alpha_t}}
	\left(
	\mathrm{x}_{1:L}^{t}
	-
	\frac{\beta_t}{\sqrt{1-\bar{\alpha}_t}}
	\epsilon_{\theta,1:L}\!\left(
	\mathrm{x}_{1:L}^{t},t,\mathrm{c}
	\right)
	\right)
	+
	\sqrt{\beta_t}\,z_{1:L},
	\qquad
	z_{1:L}\sim\mathcal{N}(0,I).
	\label{eq:reverse_sample}
\end{equation}
The model is trained by minimizing the mean-squared error (MSE) loss between the added noise sequence ${\epsilon}^t_{1:L}$ at the forward process step $t$ and the predicted noise  ${\epsilon}_{\theta,1:L}(\cdot)$
\begin{equation} 
	\mathcal{L}_{\mathrm{noise}} 
	= 
	\frac{1}{T} 
	\sum_{t=1}^{T} 
	\left\| 
	\epsilon^t_{1:L} 
	- 
	\epsilon_{\theta,1:L} 
	\left( 
	\mathrm{x}_{1:L}^{t}, 
	t, 
	{c} 
	\right) 
	\right\|^2_2.
	\label{eq:mse_loss} 
\end{equation}
Once the model learns how to remove the noise, it can generate new trajectories from pure noise. In our study, each trajectory consists of seven time-varying variables: the concentrations of CO, CO$_2$, H$_2$O, and H$_2$, the reactor temperature $T_r$, the coolant temperature $T_c$, and the feed flow rate $F$. The clean and noisy trajectories (at diffusion step $t$) are represented as $\mathrm{x}_{1:L}^{0}, \mathrm{x}_{1:L}^{t}\in\mathbb{R}^{m\times L}$, $m=7$ for our study. The conditioning variables are the heat transfer efficiency factor $f$ and the activation energy factor $f_{E_a}$, represented as $c\in\mathbb{R}^{n_c}$, $n_c=2$; see Fig. \ref{fig:diffusion_time_series}. 

To incorporate the physics loss into the reverse process, the generated inputs by the diffusion model at diffusion step $t$, $T_c^t$ and $F^t$, are provided to the WGS reaction simulator, which solves the governing laws Eq. (10)-(16). The simulator outputs physically-driven reactor temperature and species concentration trajectories that the diffusion model aims to produce if the governing laws are fully respected. At difusion step $t$, the difference between synthetic outputs $T^t_{r,\theta}$, $C^t_{i,\theta}$ and simulator outputs $T^t_{r,sim},C^t_{i,sim}$ represent the extent at which the diffusion model respects governing laws. This error will serve as a regularization to train the diffusion model
\begin{equation}
	\mathcal{L}_{\mathrm{physics}} 
	= \frac{1}{T}\sum_{t=1}^{T}\left[
\|
	T^t_{{r,sim}}-
	T^t_{r,\theta} 
	\|_2^2
	+ 
	\sum_{i=1}^{4} 
	\|
	C^t_{i,{sim}}-
	C^t_{i,\theta} 
	\|_2^2\right].
	\label{eq:physics_loss}
\end{equation}
The total loss is defined as a weighted combination of the noise prediction loss (for reproducing the raw data)  and the physics loss (for complying with physics):
\begin{equation}
	\mathcal{L}_{\mathrm{total}}
	=
	\mathcal{L}_{\mathrm{noise}}
	+
	\lambda\,
	\mathcal{L}_{\mathrm{physics}},
	\label{eq:total_loss}
\end{equation}
where $\lambda$ is a weight. Note that when simulating $T_{r,sim}^t,C_{sim,i}^t$, the inputs to the simulator, although generated by the diffusion network with parameter $\theta$, are disconnected from the computational graph (irrelevant to $\theta$).

\begin{figure}[tbh]
	\centering
	\makebox[\textwidth][c]{%
		\includegraphics[width=1\textwidth]{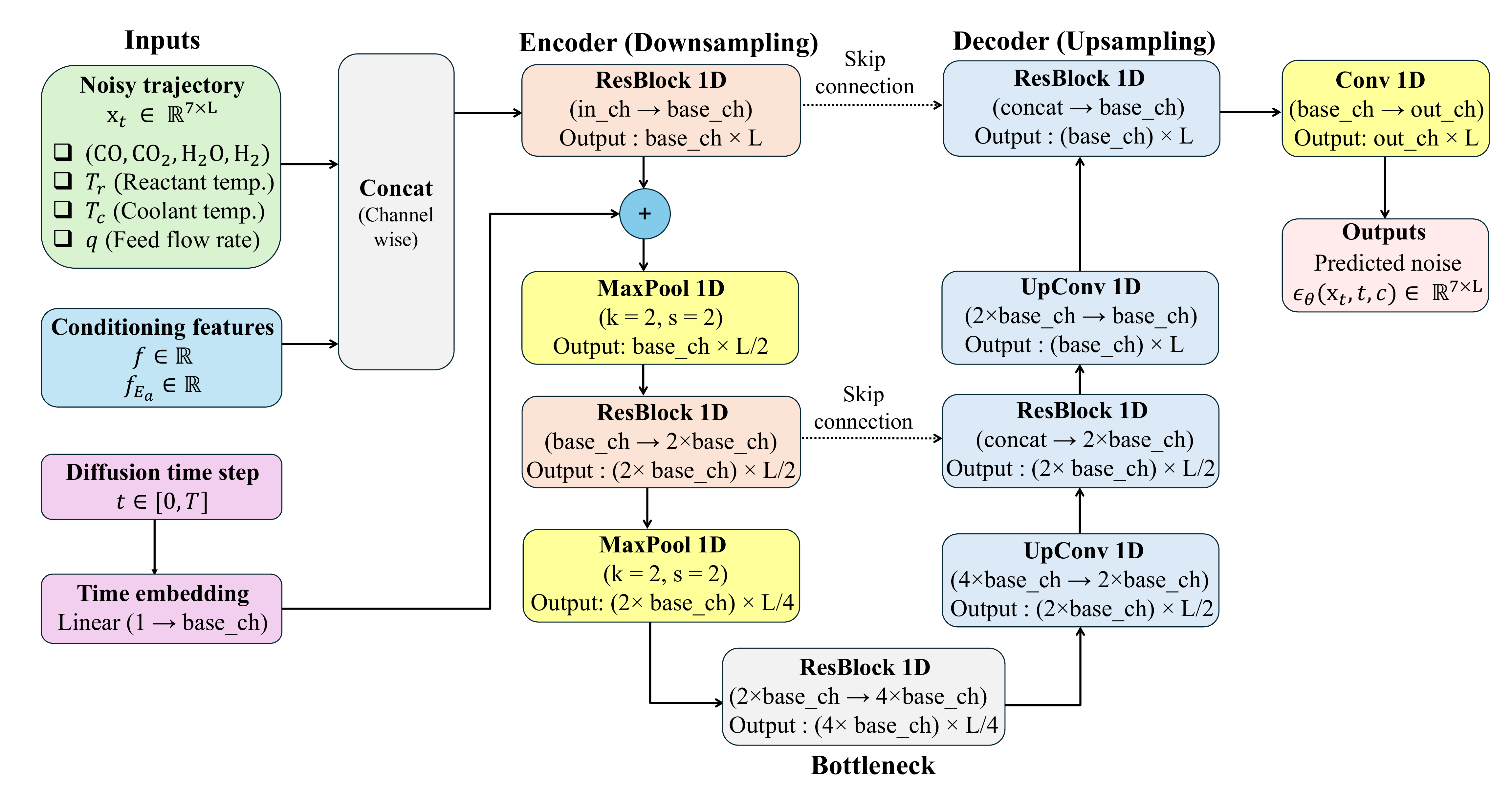}
	}
	\caption{Architecture of the proposed conditional 1D U-Net used to predict the added noise during the reverse diffusion process.}
	\label{fig:U-Net Architecture}
\end{figure}
\subsubsection{U-Net architecture}
The proposed Pg-CDDPM utilizes an 1D U-Net to predict the noise added during the forward process. The 1D U-Net employs 1D convolution operations, which are suitable for time-series applications \cite{deng2023u,kuang20191d}. The 1D U-Net consists of three main components: an encoder, a bottleneck, and a decoder \cite{ronneberger2015u}. The encoder downsamples the noisy input trajectory to extract local features, the bottleneck captures global features from the compressed noisy trajectory, and the decoder upsamples the feature maps back to the original trajectory length to predict the noise. 

\begin{figure}[tbh]
	\centering
	\makebox[\textwidth][c]{%
		\includegraphics[width=0.8\textwidth]{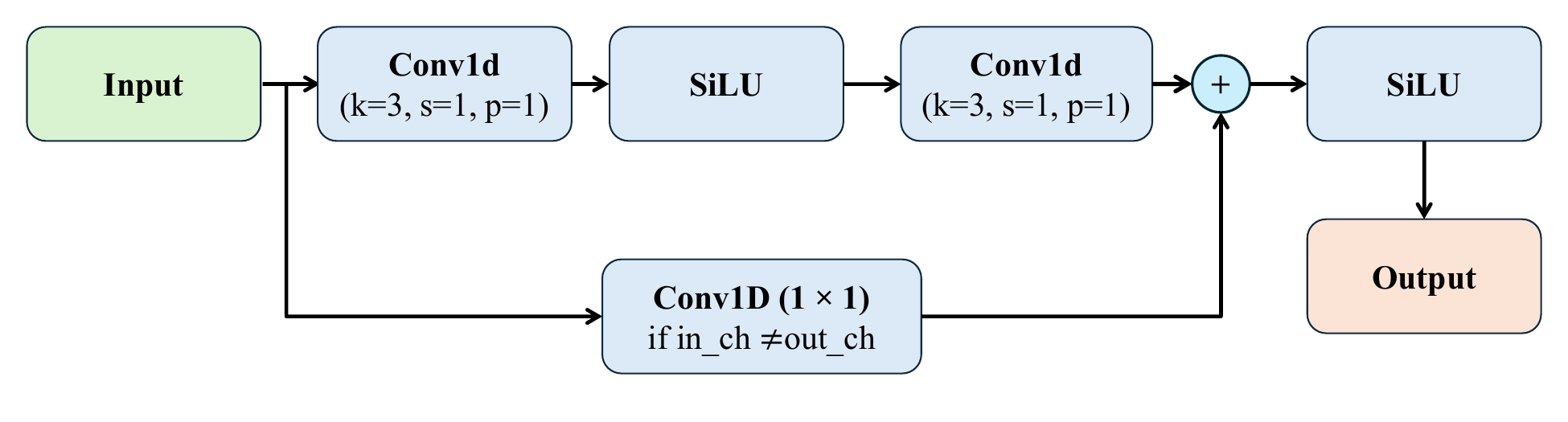}
	}
	\caption{Architecture of the ResBlock 1D used as building blocks of the diffusion model.}
	\label{fig:Resblock Architecture}
\end{figure}

As shown in Fig. \ref{fig:U-Net Architecture}, the network takes three inputs: the noisy trajectory $\mathrm{x}_{1:L}^{t}\in\mathbb{R}^{m\times L}$, the conditioning features $c\in\mathbb{R}^{n_c}$, and the diffusion time step $t$. The conditioning features are concatenated with the noisy trajectory along the channel dimension. In the early diffusion steps, only a small amount of noise is added to the input trajectory, whereas the later diffusion steps contain significantly higher noise levels. Thus, the diffusion time step is first converted into a time embedding using a linear layer and then added to the encoder features. This provides the 1D U-Net with information about the current diffusion step, allowing it to understand how much noise has been added to the input trajectory and accurately predict the noise. The encoder consists of two ResBlock 1D layers, each followed by a MaxPool1D layer to extract important features from the input trajectory. The ResBlock 1D structure has been shown in Fig. \ref{fig:Resblock Architecture}. A bottleneck ResBlock further processes the encoded features before they are passed to the decoder. The decoder consists of two UpConv1D layers and two ResBlock 1D layers that gradually reconstruct the features to the original trajectory length $L$. Skip connections are used between the encoder and decoder to retain important information from the input trajectory. Finally, a $1\times1$ Conv1D layer maps the decoder features to $m$ output channels, producing the predicted noise ${\epsilon}_{\theta,1:L}(\mathrm{x}_{1:L}^{t},t,c)\in\mathbb{R}^{m\times L}$. 

Our task-specific contributions to the 1D U-Net architecture are summarized as follows: (1) we incorporate ResNet blocks into the 1D U-Net instead of simple 1D convolutional operations. The ResNet blocks learn the residual changes, making them effective for predicting the residual noise representations in the trajectories, while improving gradient propagation during training; and (2) within each ResNet block, the SiLU activation function is employed because it is effective in capturing both positive and negative noise components. The overall architecture of the proposed 1D U-Net is shown in Table~\ref{tab:unet_architecture}.
	
	
	

\begin{table}[!ht]
	\centering
	\caption{Summary of the proposed 1D U-Net architecture.}
	\label{tab:unet_architecture}
	\renewcommand{\arraystretch}{1.2}
	\resizebox{\textwidth}{!}{
	\begin{tabular}{cccccc}
		\hline
		\textbf{Layer} &
		\textbf{Operation} &
		\textbf{Kernel} &
		\textbf{Stride} &
		\textbf{Channels (In $\rightarrow$ Out)} &
		\textbf{Trajectory (In $\rightarrow$ Out)} \\
		\hline
		
		Input &
		Concatenate inputs &
		-- &
		-- &
		$7+2 \rightarrow 9$ &
		$L \rightarrow L$ \\
		
		Resnet Block 1 &
		ResBlock 1D &
		3 &
		1 &
		$9 \rightarrow \mathrm{base\_ch}$ &
		$L \rightarrow L$ \\
		
		Maxpooling 1 &
		MaxPool1D &
		2 &
		2 &
		$\mathrm{base\_ch}\rightarrow\mathrm{base\_ch}$ &
		$L \rightarrow \frac{L}{2}$ \\
		
		Resnet Block 2 &
		ResBlock 1D &
		3 &
		1 &
		$\mathrm{base\_ch}\rightarrow 2 \times \mathrm{base\_ch}$ &
		$\frac{L}{2}\rightarrow\frac{L}{2}$ \\
		
		Maxpooling 2 &
		MaxPool1D &
		2 &
		2 &
		$2 \times \mathrm{base\_ch}\rightarrow2 \times \mathrm{base\_ch}$ &
		$\frac{L}{2}\rightarrow\frac{L}{4}$ \\
		
		Bottleneck &
		ResBlock 1D &
		3 &
		1 &
		$2 \times \mathrm{base\_ch}\rightarrow4 \times \mathrm{base\_ch}$ &
		$\frac{L}{4}\rightarrow\frac{L}{4}$ \\
		
		Upsampling 1 &
		UpConv1D &
		2 &
		2 &
		$4 \times \mathrm{base\_ch}\rightarrow2 \times \mathrm{base\_ch}$ &
		$\frac{L}{4}\rightarrow\frac{L}{2}$ \\
		
		Resnet Block 3 &
		ResBlock 1D &
		3 &
		1 &
		$4 \times \mathrm{base\_ch}\rightarrow2 \times \mathrm{base\_ch}$ &
		$\frac{L}{2}\rightarrow\frac{L}{2}$ \\
		
		Upsampling 2 &
		UpConv1D &
		2 &
		2 &
		$2 \times \mathrm{base\_ch}\rightarrow\mathrm{base\_ch}$ &
		$\frac{L}{2}\rightarrow L$ \\
		
		Resnet Block 4 &
		ResBlock 1D &
		3 &
		1 &
		$2 \times \mathrm{base\_ch}\rightarrow\mathrm{base\_ch}$ &
		$L\rightarrow L$ \\
		
		Output &
		Conv1D &
		1 &
		1 &
		$\mathrm{base\_ch}\rightarrow7$ &
		$L\rightarrow L$ \\
		
		\hline
	\end{tabular}}
\end{table}

\subsubsection{Training and sampling algorithms}
As noted before, the inputs generated by the diffusion model are detached from the computational graph when being passed to the WGS reaction simulator. The physics loss is computed by comparing simulated and predicted reactor output variables, and backpropagation is performed only through the reactor output variables predicted by the diffusion model. Also, the generated coolant temperature $T_c$ and the feed flow rate $F$ are clipped to proper ranges: $350$--$500~\mathrm{K}$ and $70$--$150~\mathrm{L\,min^{-1}}$. This ensures that the diffusion-generated inputs remain within feasible operating limits before being passed to the simulator. 
Specifically, the selected coolant temperature limits can maintain the reactor temperature at approximately $300$--$450^{\circ}\mathrm{C}$ ($573$--$723~\mathrm{K}$), which is suitable for high-temperature shift operation and prevents unrealistic operating conditions. Similarly, constraining the feed flow rate ensures a realistic reactor space time $\tau=V/F\in [0.67~\text{min}, 1.43~\text{min} ]$ where $V=100~ \mathrm{L}$. These constraints can assist the diffusion model to produce plausible reactor dynamics and generating stable trajectories.

\begin{algorithm}[!ht]
	\caption{Pg-CDDPM Training}
	\label{alg:training}
	\begin{algorithmic}[1]
		
		\STATE Standardize each trajectory using training-data mean, standard deviation: $\mathrm{x}_0=\frac{\mathrm{x}_{0}-\mu}{\sigma}$.
		
		\REPEAT
		
		\STATE Sample a mini-batch $(\mathbf{x}_0,\mathbf{c})$,  where $\mathbf{x}_0\in\mathbb{R}^{N\times m\times L}$ is the standardized trajectory tensor, $\mathbf{c}\in\mathbb{R}^{N\times n_c}$ is the conditioning tensor, and $N$ is the number of trajectories in the mini-batch.
		
		\STATE Sample one diffusion step out of $[t_1,t_2,\ldots,t_N]$, $t_i\sim U\{1,\ldots,T\}$ for each trajectory in the mini-batch, and store sampled time steps into $\mathbf{t}\in\mathbb{R}^{N}$.
		
		\STATE Sample a Gaussian noise tensor $\epsilon\sim\mathcal{N}(0,\mathrm{I})$, $\epsilon\in\mathbb{R}^{N\times m\times L}$.
		
		\STATE Generate the noisy trajectories $\mathbf{x}^{\mathbf{t}}=\sqrt{\bar{\mathbf{\alpha}}_\mathbf{t}}\otimes\mathbf{x}^{0}+\sqrt{1-\bar{\mathbf{\alpha}}_\mathbf{t}}\otimes\epsilon$, where $\bar{\alpha}_{\mathbf{t}}:=[\bar{\alpha}_{t_1},\ldots,\bar{\alpha}_{t_N}]^\top\in\mathbb{R}^{N}$ stacks the $\bar{a}_t$ as in \eqref{eq:alpha} for each sample in the batch, and $\otimes$ denotes sample-wise product. 
		
		\STATE Predict the noise tensor $\epsilon_{\theta}\left(\mathbf{x}^{\mathbf{t}},\mathbf{t},\mathbf{c}\right)\in\mathbb{R}^{N\times m\times L}$ using the 1D U-Net.
		
		\STATE Compute the diffusion noise loss $\mathcal{L}_{\mathrm{noise}}$ by applying \eqref{eq:mse_loss} across all samples in the mini-batch.
		
		\STATE Estimate the clean trajectory using the predicted noise $\hat{\mathbf{x}}^{0}=\frac{\mathbf{x}^{\mathbf{t}}-\sqrt{1-\bar{\alpha}_\mathbf{t}}\otimes\epsilon_{\theta}\left(\mathbf{x}^{\mathbf{t}},\mathbf{t},\mathbf{c}\right)}{\sqrt{\bar{\alpha}_\mathbf{t}}}$.

		\STATE Split the predicted clean trajectory into outputs $\hat{\mathbf{s}}_{\theta}^{sd}=[\hat{T}_{r,\theta},\hat{C}_{i,\theta}]$ and control inputs $\hat{\mathbf{u}}_{\theta}^{sd}=[\hat{T}_{c,\theta},\hat{F}_{\theta}]$, where the superscript ``sd'' means standardized. 
		
		\STATE Inverse-standardize the predicted control inputs to recover their physical values $\hat{\mathbf{u}}_{\theta}$.
		
		\STATE Detach the generated control inputs (gradient free):
		$\hat{\mathbf{u}}:=[\hat{T}_c,\hat{F}]
		\leftarrow
		\mathrm{detach}(\hat{\mathbf{u}}_{\theta})$ from the computation graph. 
		
		\STATE Clip the control inputs $\hat{T}_c$ and $\hat{F}$ to the ranges of $350$--$500~\mathrm{K}$ and $70$--$150~\mathrm{L\,min^{-1}}$.

		\STATE Randomly initialize the reactor using initial and feed conditions as in Table \ref{tab:initial_feed_conditions}.
		
		\STATE Simulate the CSTR using
		$\hat{T}_c,\hat{F}$ and above initial conditions to obtain simulated outputs, followed by standardization to obtain $[\hat{T}^{sd}_{r,sim},\hat{C}^{sd}_{i,sim}]$.
		
		\STATE Compute the physics loss $\mathcal{L}_{\mathrm{physics}}$ as in \eqref{eq:physics_loss} and total loss $\mathcal{L}_{\mathrm{total}}$ as in \eqref{eq:total_loss}.

		\STATE Update the network parameters $\theta$ via backpropogation of $\mathcal{L}_{\mathrm{total}}$ using the Adam optimizer.
		
		\UNTIL{$\theta$ converges to $\theta^*$}
	\end{algorithmic}
\end{algorithm}

Algorithm~1 summarizes the training procedure of the proposed Pg-CDDPM. It describes the forward process, noise prediction, reconstruction of clean trajectories, computation of physics loss using the CSTR simulator, and optimization of model parameters. Algorithm~2 presents the conditional sampling procedure, where the trained model gradually denoises an initial noise sample through the reverse diffusion process to generate high-quality synthetic reaction data under specified operating conditions.

\begin{algorithm}[H]
	\caption{Conditional Sampling Procedure}
	\label{alg:sampling}
	\begin{algorithmic}[1]
		
	\STATE Sample the initial Gaussian noise $\mathrm{x}_{1:L}^T\sim\mathcal{N}(\mathrm{0},\mathrm{I})$.
		
	\FOR{$t=T,T-1,\ldots,1$}
		
	\STATE Predict the Gaussian noise ${{\epsilon_{\theta^*,1:L}(\mathrm{x}_{1:L}^t,t,c)}}$
	
	\STATE Compute the reverse-diffusion mean using the predicted noise \\
	$\mu_{\theta^*,1:L}=\frac{1}{\sqrt{\alpha_t}}\left(\mathrm{x}_{1:L}^t-\frac{\beta_t}{\sqrt{1-\bar{\alpha}_t}}\epsilon_{\theta^*,1:L}\left(\mathrm{x}_{1:L}^t,t,c\right)\right)$.
	
	\IF{$t>1$}
	
	\STATE Sample $z_{1:L}\sim\mathcal{N}(\mathrm{0},\mathrm{I})$.
	
	\STATE Sample the previous trajectory $\mathrm{x}_{1:L}^{t-1}=\mu_{\theta^*,1:L}+\sqrt{\beta_t}z_{1:L}$.
	
	\ENDIF
	
	\STATE Set $\mathrm{x}_{1:L}^0={\mu}_{\theta^*,1:L}$ when $t=1$
		
	\ENDFOR
		
	\STATE \textbf{return} $\mathrm{x}_{1:L}^0$
	
	\STATE Inverse-standardize $\mathrm{x}_{1:L}^0$ using training data mean and standard deviation to nominal ranges.

	\end{algorithmic}
\end{algorithm}

\begin{figure}[htb]
	\centering
	\makebox[\textwidth][c]{%
		\includegraphics[width=0.9\textwidth]{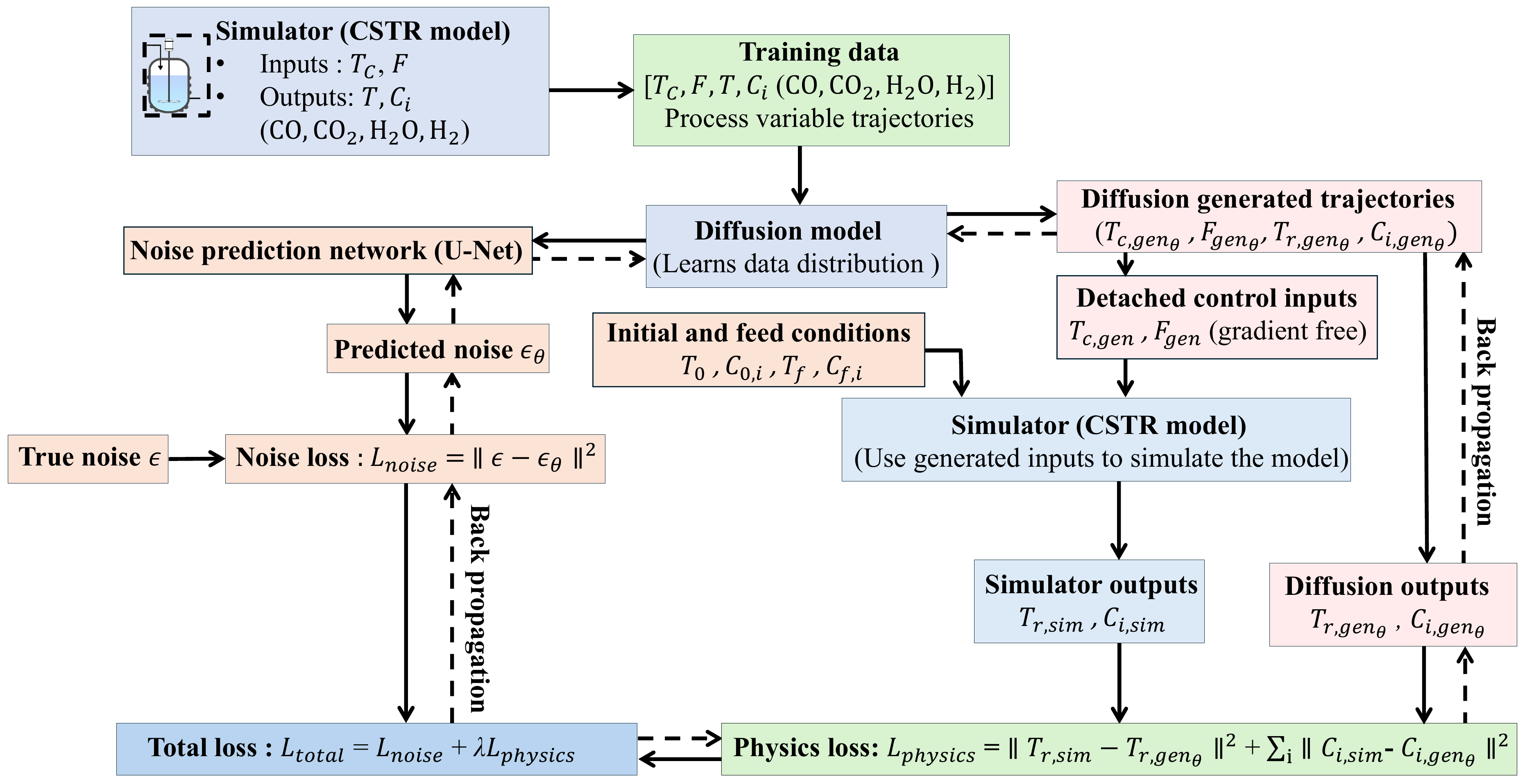}
	}
	\caption{The flow chart for training the proposed Pg-CDDPM to generate high-fidelity synthetic profiles for the WGS reaction.}
	\label{fig:Overall Block Diagram}
\end{figure}

The detailed training workflow of the proposed Pg-CDDPM is shown in Fig. ~\ref{fig:Overall Block Diagram}. First, the diffusion model learns the distribution of training trajectories by predicting the added noise using the U-Net. The predicted noise is then adopted to reconstruct the clean trajectories. Next, the reconstructed trajectories are separated into generated input and output variables. The generated inputs are detached from the gradient computation and passed to the CSTR simulator to obtain corresponding physics-based trajectories. The simulator outputs are then compared with the diffusion-generated state trajectories to calculate the physics loss. Finally, the noise and physics losses are combined to obtain the total training loss to train the diffusion model.

\subsection{Hazard Score Index}
\begin{figure}[!ht]
 	\centering
 	
 	\begin{minipage}[t]{0.45\textwidth}
 		\centering
 		\vspace{0pt}
 		\includegraphics[
 		height=3.5cm
 		]{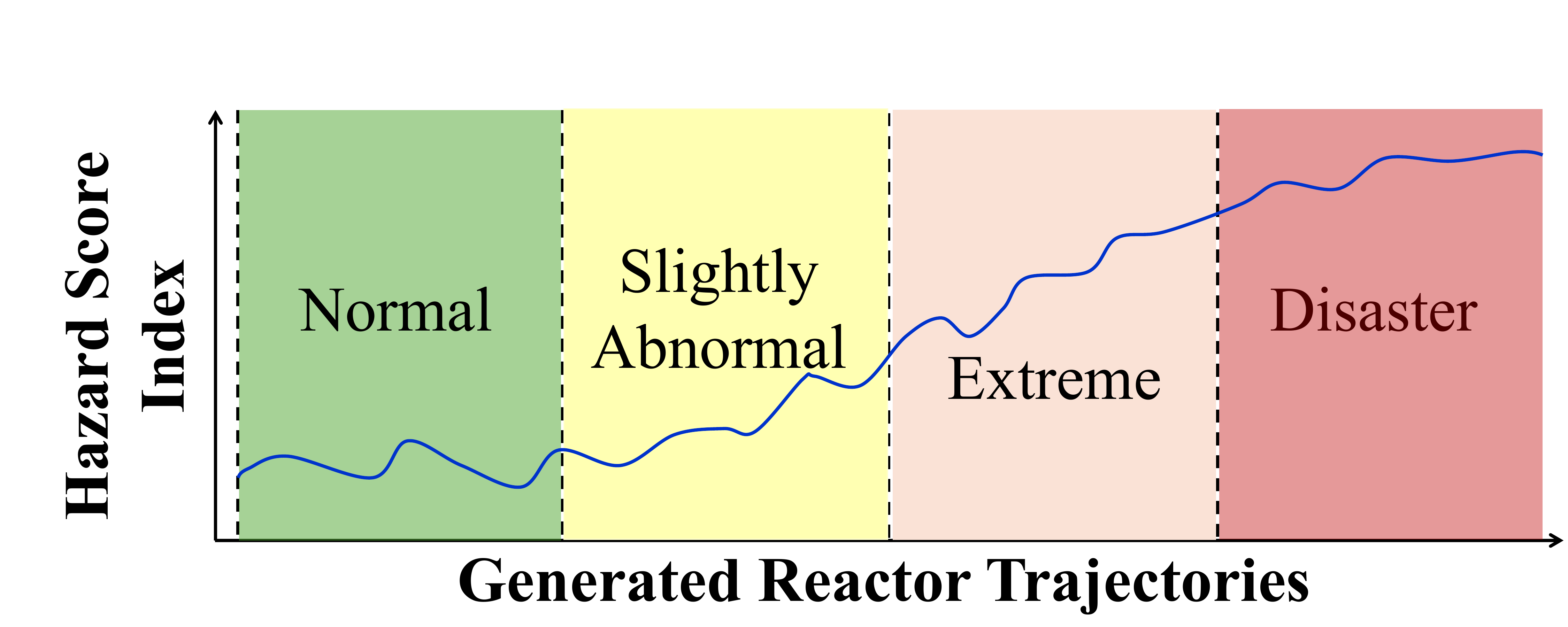}\\
 		\textbf{(a)} Hazard score index
 	\end{minipage}
 	\hfill
 	\begin{minipage}[t]{0.45\textwidth}
 		\centering
 		\vspace{0pt}
 		\includegraphics[
 		height=3.5cm
 		]{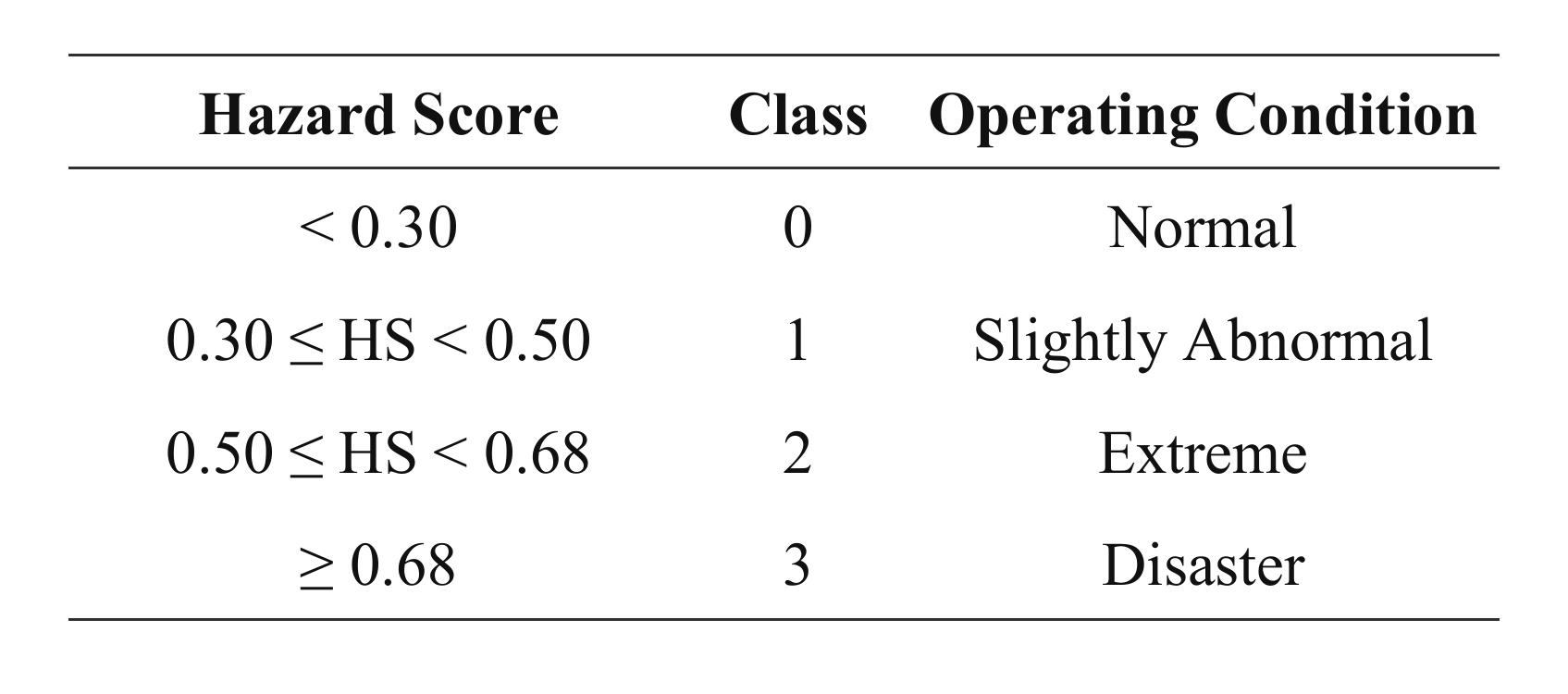}\\
 		\textbf{(b)} Hazard classification
 	\end{minipage}
 	
 	
\begin{minipage}[tb]{0.45\textwidth}
	\centering
	\vspace{0pt}
	\includegraphics[
	height=4.5cm
	]{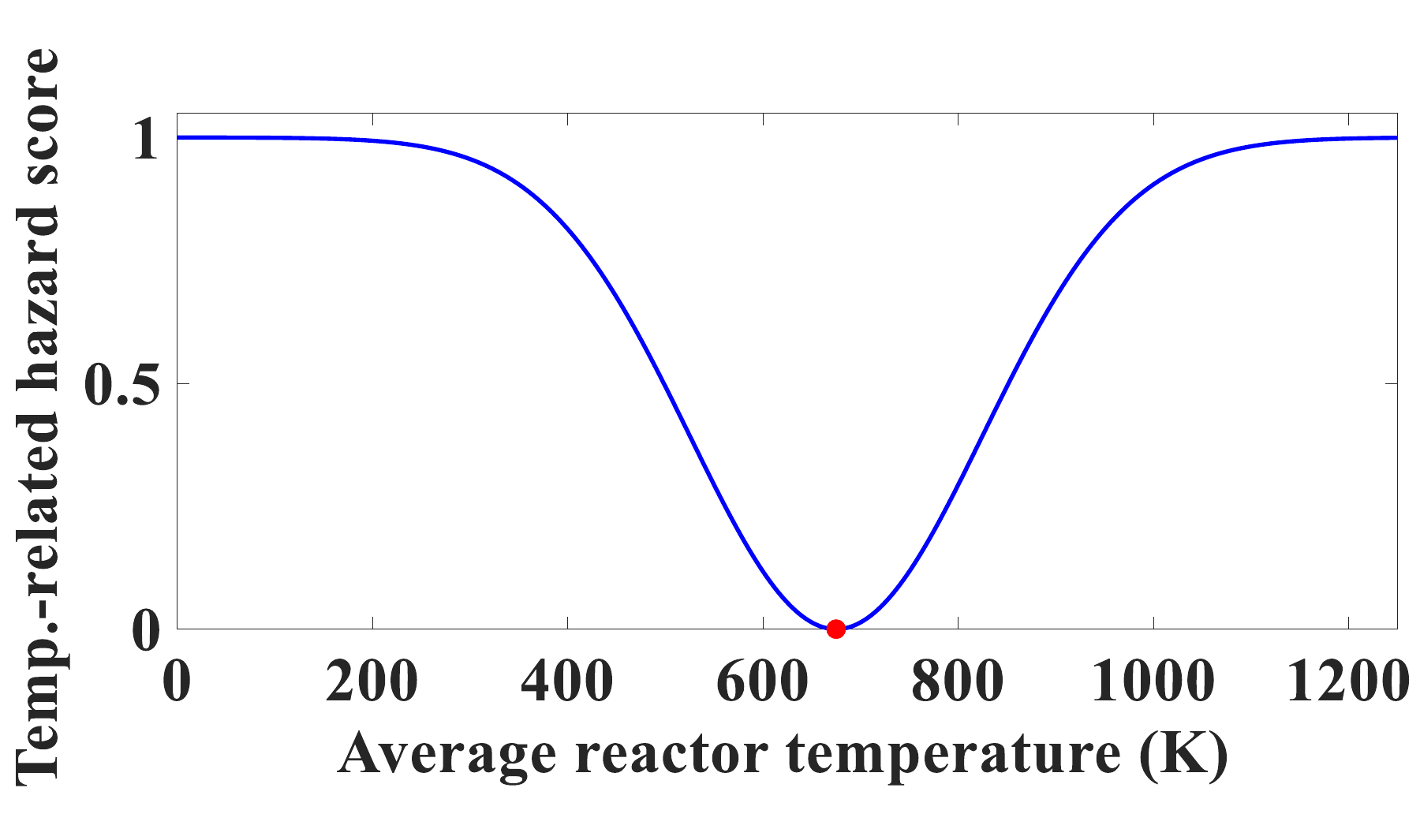}\\
	\textbf{(c)} Temperature-related hazard score
\end{minipage}
\hfill
\begin{minipage}[tb]{0.45\textwidth}
	\centering
	\vspace{0pt}
	\includegraphics[
	height=4.5cm
	]{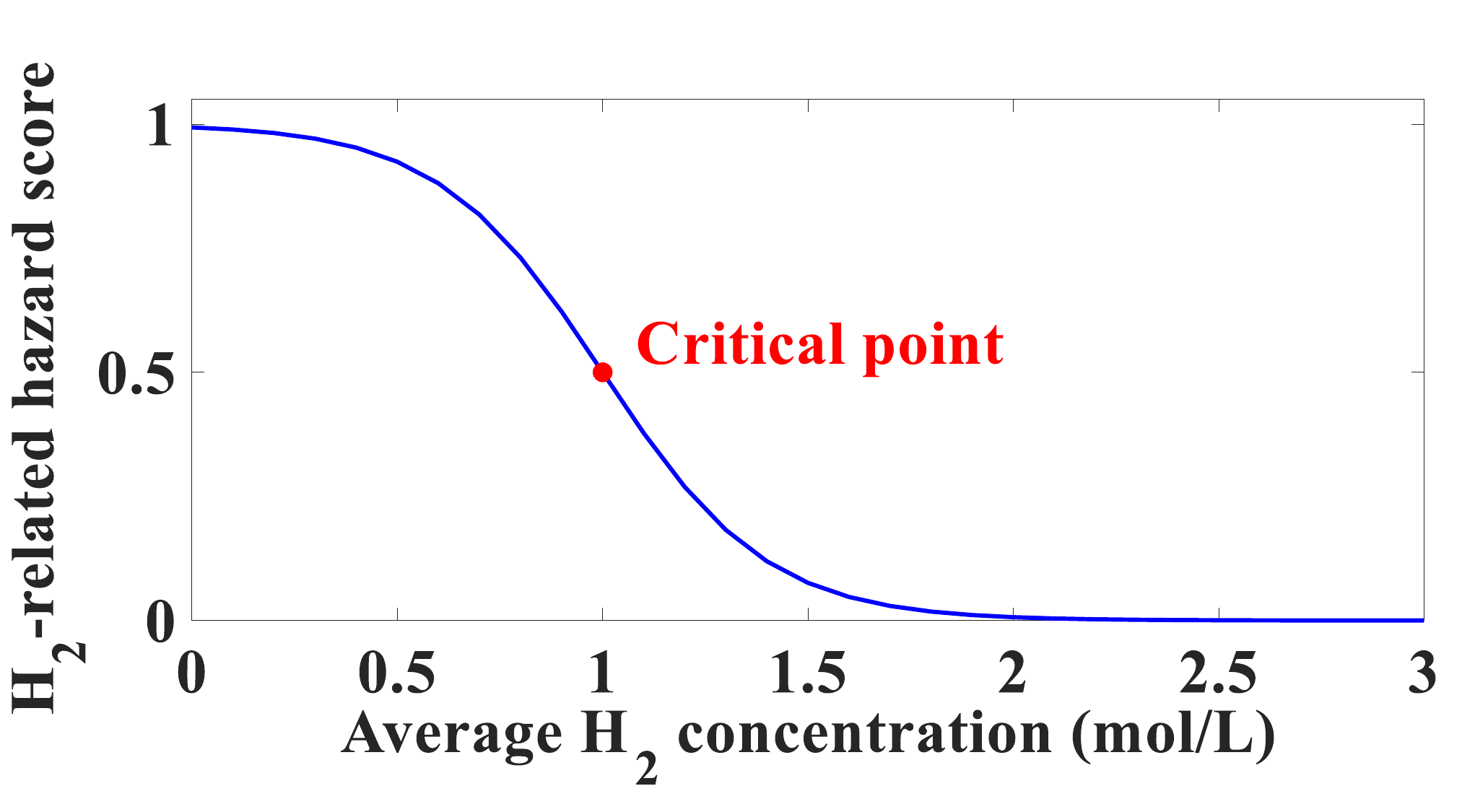}\\
	\textbf{(d)} Hydrogen-related hazard score
\end{minipage}

\caption{Hazard-based classification of generated reactor trajectories,
	classification thresholds, and temperature- and hydrogen-related
	contributions to the hazard score.}
\label{fig:hazard_score_framework}
\end{figure}

The established physics-guided diffusion model above can assist to generate synthetic operating profiles that are physically consistent. This is particularly useful for the monitoring and diagnosis of rare-event scenarios where data scarcity is a predominant issue. For effective fault diagnosis, these operating profiles must be categorized according to the risk severity of the reactor condition. This severity classification converts complex multivariate trajectories into clear hazard levels that can be used to train and evaluate diagnostic models. To quantitatively reflect the operation status of the WGS reaction, we present a comprehensive hazard score as a health indicator for this process. Specifically, this hazard score combines the effects of reactor temperature, hydrogen concentration, heat transfer degradation, and catalyst degradation:
\begin{equation}
	\begin{aligned}
		\mathrm{Hazard\ Score}
		={}&
		0.4\frac{1}
		{1+\exp\!\left[0.5\left(\bar{C}_{\mathrm{H_2}}
			-C_{\mathrm{H_2,crit}}\right)\right]}
		\\
		&+0.2\left[
		1-\exp\!\left(
		-\frac{\left(\bar{T_r}-T_0\right)^2}{2\sigma_T^2}
		\right)
		\right]
		+0.2(1-f)
		+0.2(f_{E_a}-1),
	\end{aligned}
	\label{eq:hazard_score}
\end{equation}
where $\bar{C}_{\mathrm{H_2}}$ and $C_{\mathrm{H_2,crit}}$ denote the average and critical hydrogen concentrations, while $\bar{T_r}$, $T_0$, and $\sigma_T$ denote the average reactor temperature, optimal operating temperature, and temperature spread around $T_0$, respectively. In this study, we consider $C_{\mathrm{H_2,crit}}=1~\mathrm{mol/L}$, $T_0=675~\mathrm{K}$ ($402^{\circ}\mathrm{C}$), and $\sigma_T=150~\mathrm{K}$ (i.e., a temperature spread of $150^{\circ}\mathrm{C}$).
First, the reactor temperature and hydrogen concentration are averaged over the trajectory length $L$. The hazard score is formulated such that both extremely low and high temperatures increase the hazard level. Low temperatures result in slow reaction kinetics, whereas high temperatures may lead to thermal runaway. The parameter $\sigma_T$ controls the width of the acceptable temperature range, beyond which the hazard score increases rapidly. Similarly, the hazard score rises sharply when $\bar{C}_{\mathrm{H_2}}$ falls below $C_{\mathrm{H_2,crit}}$. The dependence of the hazard score on $\bar{T}$ and $\bar{C}_{\mathrm{H_2}}$ is shown in Figs.~\ref{fig:hazard_score_framework}(c) and \ref{fig:hazard_score_framework}(d), respectively. The factors $f$ and $f_{E_a}$ account for fouling and catalyst degradation. Thus, higher scores represent more severe degradation and hazardous operating conditions. Each trajectory will be assigned to one of the four hazard classes based on its corresponding hazard score index shown in Fig.~\ref{fig:hazard_score_framework}(b). Fig.~\ref{fig:hazard_score_framework}(a) shows the increase in hazard score from normal to disaster conditions.

\subsection{Hyperparameter Selection}
The proposed Pg-CDDPM is implemented using the PyTorch framework and trained using the Adam optimizer. Table~\ref{tab:training_parameters} summarizes the network configuration, training hyperparameters, and computational hardware used in this study. For the trajectory severity classification, we used the GRU classifier. Table~\ref{tab:gru_hyperparameters} summarizes the hyperparameters used in the GRU classifier.

\begin{table}[tbh]
	\centering
	\caption{Hyperparameters used for training the proposed Pg-CDDPM.}
	\label{tab:training_parameters}
	
	\begingroup
	\small
	\renewcommand{\arraystretch}{1.0}
	\begin{tabularx}{\textwidth}{@{}X l X l@{}}
		\hline
		\textbf{Parameter} & \textbf{Value} &
		\textbf{Parameter} & \textbf{Value} \\
		\hline
		
		Deep-learning framework & PyTorch
		& Beta schedule & Linear \\
		
		Optimizer & Adam
		& $\beta_{\mathrm{start}}$ & $10^{-4}$ \\
		
		Learning rate & $10^{-4}$
		& $\beta_{\mathrm{end}}$ & $10^{-2}$ \\
		
		Loss function & MSE
		& Input channels & 9 \\
		
		Batch size & 5
		& Output channels & 7 \\
		
		Number of epochs & 6000
		& Base channels & 64 \\
		
		Random seed & 42
		& Activation function & SiLU \\
		
		Diffusion time steps ($T$) & 1000
		& Physics-loss weight ($\lambda$) & 0.3 \\
		
		& & GPU & NVIDIA RTX 4000 \\
		
		\hline
	\end{tabularx}
	\endgroup
\end{table}

\begin{table}[tbh]
	\centering
	\caption{Training hyperparameters and architecture of the GRU classifier.}
	\label{tab:gru_hyperparameters}
	
	\begingroup
	\small
	\renewcommand{\arraystretch}{1.0}
	\begin{tabularx}{\textwidth}{@{}
			>{\raggedright\arraybackslash}X
			>{\raggedright\arraybackslash}X
			>{\raggedright\arraybackslash}X
			l@{}}
		\hline
		\textbf{Parameter} & \textbf{Value} &
		\textbf{Parameter} & \textbf{Value} \\
		\hline
		
		Deep-learning framework & PyTorch
		& Fully connected activation & ReLU \\
		
		Sequence length & 100 time steps
		& Number of output classes & 4 \\
		
		Number of input features & 7
		& Loss function & Cross-entropy \\
		
		Input features
		& CO, H$_2$O, CO$_2$, H$_2$, $T$, $T_c$, $q$
		& Optimizer & Adam \\
		
		GRU hidden size & 64
		& Learning rate & $10^{-3}$ \\
		
		Number of GRU layers & 8
		& Batch size & 50 \\
		
		GRU dropout rate & 0.2
		& Number of epochs & 300 \\
		
		Fully connected architecture & $64\rightarrow64\rightarrow4$
		& Random seed & 42 \\
		
		\hline
	\end{tabularx}
	\endgroup
\end{table}
\section{Results and Discussion}
This section will evaluate: (i) the performance of the proposed Pg-CDDPM method on synthesizing operating trajectories for the WGS reaction; and (ii) the critical role that synthetic data plays in assisting the monitoring and diagnosis of rare-event cases. First, the proposed diffusion model is trained using ground-truth trajectory data generated by the reactor simulator. After training, the model is used to produce synthetic profiles for different operating conditions as in Tables~5--7, particularly the disaster case where ground-truth training data are completely absent (an extrapolation problem). Physical consistency and closeness to ground-truth data are the main criteria to assess the quality of synthetic data. Further, the synthetic data are integrated with the raw data to address the data scarcity issue, especially for rare-event scenarios. The mixed datasets will then be leveraged to diagnose different hazard classes, by training a classifier that reliably classify these classes for any future operating data. The classification performance is used to assess the effectiveness of the entire synthesis-and-diagnosis framework. 

\subsection{Data Synthesis of the Proposed Pg-CDDPM}
\subsubsection{Training trajectory generation}
To construct the training dataset for the Pg-CDDPM, we used the nonlinear CSTR model described in Section 3.1 to generate dynamic trajectories.
Each trajectory consists of 100 time steps. The initial reactor conditions were randomly selected within the ranges listed in Table~\ref{tab:initial_feed_conditions} to produce diverse training trajectories. The inputs were varied within the ranges given in Table~\ref{tab:manipulated_variables} using randomly generated piecewise-constant input profiles with an input holding time of 10 simulation steps. The feed conditions used to generate each trajectory are listed in Table~\ref{tab:initial_feed_conditions} . Conditioning features were varied within the ranges listed in Table~\ref{tab:process_parameters} to create different operating conditions. 
The ranges of $f$ and $f_{E_a}$ used to generate the training trajectories are summarized in Table~\ref{tab:trajectory_ranges}.


\subsubsection{Ablation studies for interpolation and extrapolation}
We have conducted ablation studies to evaluate the contribution of the physics-guided loss to the trajectory generation performance of the proposed model. The CDDPM was trained and evaluated both without and with the physics-guided loss while keeping the remaining model architecture and training settings unchanged.


\begin{table}[!ht]
	\centering
	\small
	\caption{Ranges of input variables used for generating the input profiles.}
	\label{tab:manipulated_variables}
	\begin{tabular}{lll}
		\hline
		\textbf{Input Variable} & \textbf{Symbol} & \textbf{Range} \\
		\hline
		Coolant temperature & $T_c$ & $350$--$500~\mathrm{K}$ \\
		Feed flow rate & $F$ & $70$--$150~\mathrm{L\,min^{-1}}$ \\
		Input holding time & $N_{\mathrm{hold}}$ & $10$ time steps \\
		\hline
	\end{tabular}
\end{table}

\begin{table}[!ht]
	\centering
	\caption{Initial reactor and feed conditions used for generating the training trajectories. Concentrations are in $\mathrm{mol\,L^{-1}}$ and temperatures are in $\mathrm{K}$.}
	\label{tab:initial_feed_conditions}
	
	\begingroup
	\small
	\renewcommand{\arraystretch}{1.0}
	\begin{tabularx}{\textwidth}{@{}
			>{\raggedright\arraybackslash}X
			c
			c
			>{\raggedright\arraybackslash}X
			c
			c@{}}
		\hline
		\multicolumn{3}{c}{\textbf{Initial reactor conditions}} &
		\multicolumn{3}{c}{\textbf{Feed conditions}} \\
		\cline{1-3}\cline{4-6}
		\textbf{Parameter} & \textbf{Symbol} & \textbf{Value} &
		\textbf{Parameter} & \textbf{Symbol} & \textbf{Value} \\
		\hline
		
		Initial CO concentration & $C_{\mathrm{CO},0}$ & $1.5$--$2.0$ &
		Feed CO concentration & $C_{f,\mathrm{CO}}$ & $1.5$ \\
		
		Initial H$_2$O concentration & $C_{\mathrm{H_2O},0}$ & $3.0$--$3.5$ &
		Feed H$_2$O concentration & $C_{f,\mathrm{H_2O}}$ & $3.0$ \\
		
		Initial CO$_2$ concentration & $C_{\mathrm{CO_2},0}$ & $0.2$ &
		Feed CO$_2$ concentration & $C_{f,\mathrm{CO_2}}$ & $0.0$ \\
		
		Initial H$_2$ concentration & $C_{\mathrm{H_2},0}$ & $0.2$ &
		Feed H$_2$ concentration & $C_{f,\mathrm{H_2}}$ & $0.0$ \\
		
		Initial reactor temperature & $T_0$ & $500$--$550$ &
		Feed temperature & $T_f$ & $550$ \\
		\hline
	\end{tabularx}
	\endgroup
\end{table}

\begin{table}[!ht]
	\centering
	\caption{Process parameters used in the WGS reactor model.}
	\label{tab:process_parameters}
	
	\begingroup
	\small
	\setlength{\tabcolsep}{3pt}
	\renewcommand{\arraystretch}{1.1}
	\resizebox{\textwidth}{!}{%
		\begin{tabular}{@{}lll lll@{}}
			\hline
			\textbf{Parameter} & \textbf{Symbol} & \textbf{Value} &
			\textbf{Parameter} & \textbf{Symbol} & \textbf{Value} \\
			\hline
			Reactor volume & $V$ & $100~\mathrm{L}$ &
			Base overall heat transfer coefficient & $UA$ & $5.0\cdot10^{4}$ \\
			
			Fluid density & $\rho$ & $1000~\mathrm{kg\,m^{-3}}$ &
			Heat transfer efficiency factor & $f$ & $0.6$--$1.0$ \\
			
			Heat capacity & $C_p$ & $0.239~\mathrm{kJ\,kg^{-1}\,K^{-1}}$ &
			Effective heat transfer coefficient & $UA_{\mathrm{eff}}$ & $UA\cdot f$ \\
			
			Pre-exponential factor & $k_0$ & $10^{10}$ &
			Base activation energy & $E_a$ & $7.2\cdot10^{4}~\mathrm{J\,mol^{-1}}$ \\
			
			Universal gas constant & $R$ & $8.314~\mathrm{J\,mol^{-1}\,K^{-1}}$ &
			Activation energy factor & $f_{E_a}$ & $1.0$--$1.5$ \\
			
			Heat of reaction & $\Delta H_r$ & $-4.1\cdot10^{4}~\mathrm{J\,mol^{-1}}$ &
			Effective activation energy & $E_{a,\mathrm{eff}}$ & $E_a\cdot f_{E_a}$ \\
			\hline
		\end{tabular}%
	}
	\endgroup
\end{table}

\begin{table}[!ht]
	\centering
	\small
	\caption{Ranges of degradation factors used to generate the training trajectories.}
	\label{tab:trajectory_ranges}
	\begin{tabular}{lll}
		\hline
		\textbf{Trajectory No.} & \textbf{Heat Transfer Efficiency Factor $f$} & \textbf{Activation Energy Factor $f_{E_a}$} \\
		\hline
		1--30    & 0.6--1.0 & 1.0--1.2 \\
		31--60   & 0.6--1.0 & 1.2--1.3 \\
		61--90   & 0.6--1.0 & 1.3--1.4 \\
		91--120  & 0.6--1.0 & 1.4--1.45 \\
		121--130 & 0.6--1.0 & 1.45--1.5 \\
		\hline
	\end{tabular}
\end{table}

\begin{figure}[!ht]
	\centering
	\makebox[\textwidth][c]{%
		\includegraphics[width=1\textwidth]{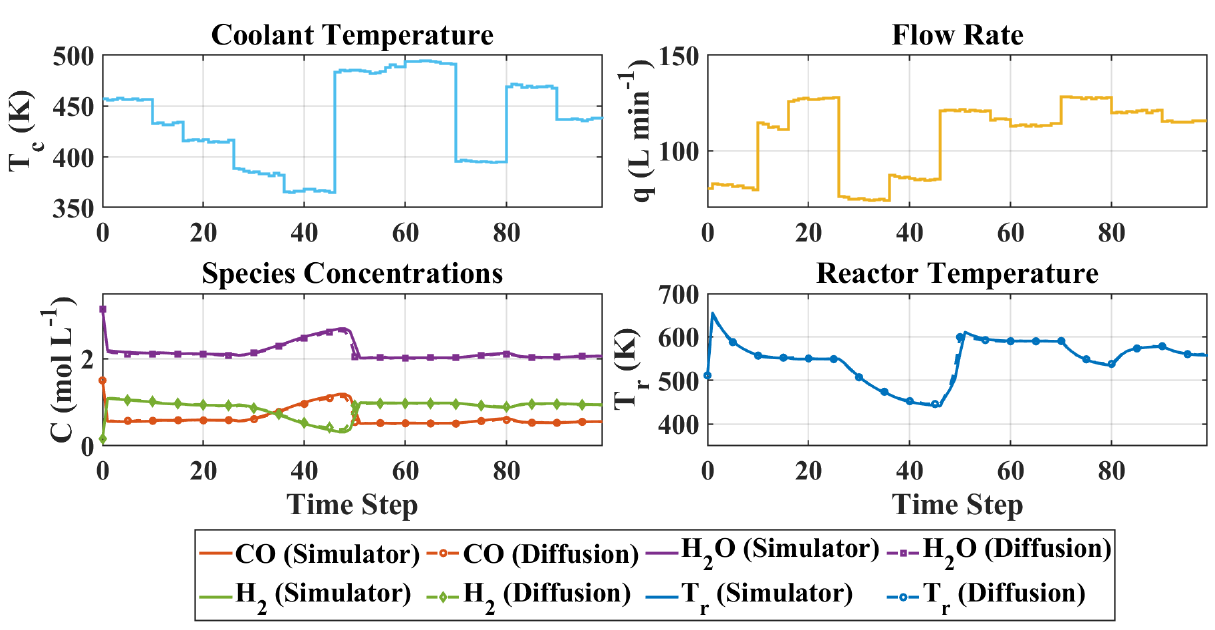}
	}
	\caption{Comparison between synthetic and ground-truth reaction profiles (interpolation) for heat transfer efficiency $f=0.72$ and activation energy $f_{E_a}=1.30$ \textit{without} the physics-guided loss.}
	\label{fig:0.72_1.30_without_physics}
\end{figure}

\begin{figure}[!ht]
	\centering
	\makebox[\textwidth][c]{%
		\includegraphics[width=1\textwidth]{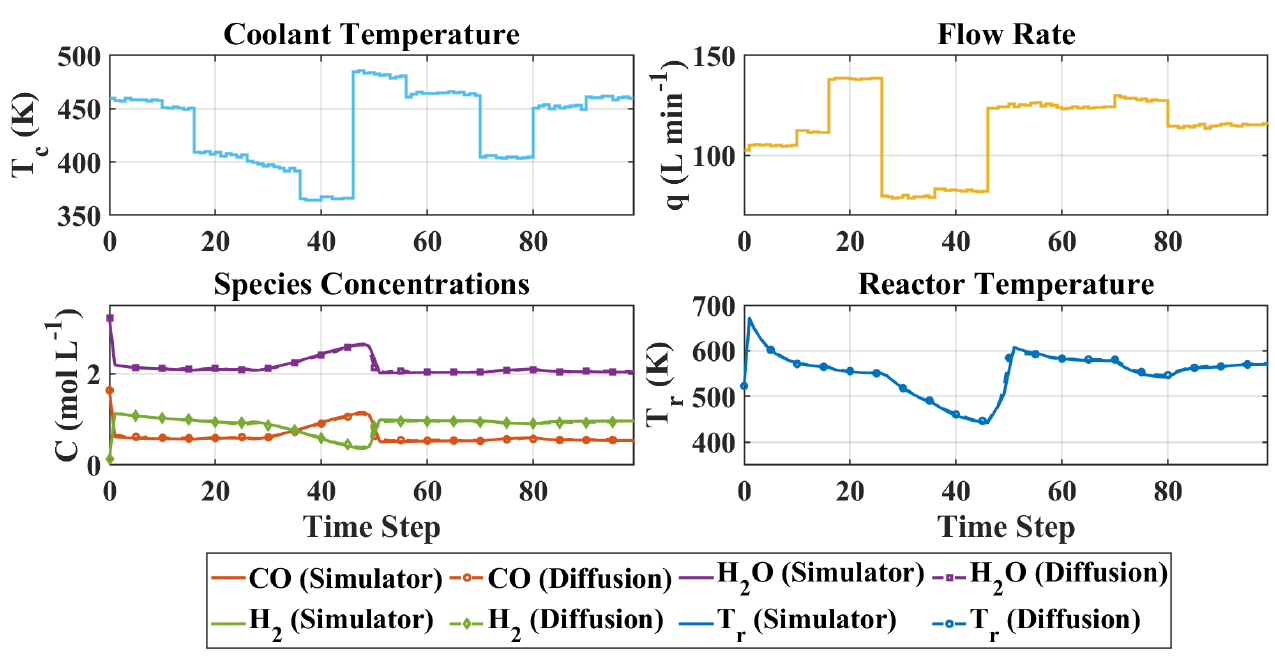}
	}
	\caption{Comparison between synthetic and ground-truth reaction profiles (interpolation) for heat transfer efficiency $f=0.72$ and activation energy $f_{E_a}=1.30$ \textit{with} the proposed Pg-CDDPM.}
	\label{fig:0.72_1.30_with_physics}
\end{figure}

Figs.~\ref{fig:0.72_1.30_without_physics} and \ref{fig:0.72_1.30_with_physics} present a representative interpolation case ($f=0.72$, $f_{E_a}=1.30$) without and with the physics-guided loss, respectively. In both cases, the generated trajectories closely agree with the simulated reactor temperature and the concentrations of CO, H$_2$O, and H$_2$ throughout the entire time horizon. Without the physics-guided loss, the temperature RMSE is approximately 5 K, while the concentration RMSEs are 0.0368, 0.0377, and 0.0348~mol L$^{-1}$ for CO, H$_2$O, and H$_2$, respectively. After incorporating the physics-guided loss, these values slightly decrease to 4 K and 0.0283, 0.0264, and 0.0267~mol L$^{-1}$, respectively. Table~\ref{tab:interpolation_ablation_rmse} summarizes the RMSE comparison for this interpolation case. These results indicate that the proposed model accurately captures the nonlinear reactor dynamics within the interpolation region.

\begin{figure}[thb]
	\centering
	\makebox[\textwidth][c]{%
		\includegraphics[width=1\textwidth]{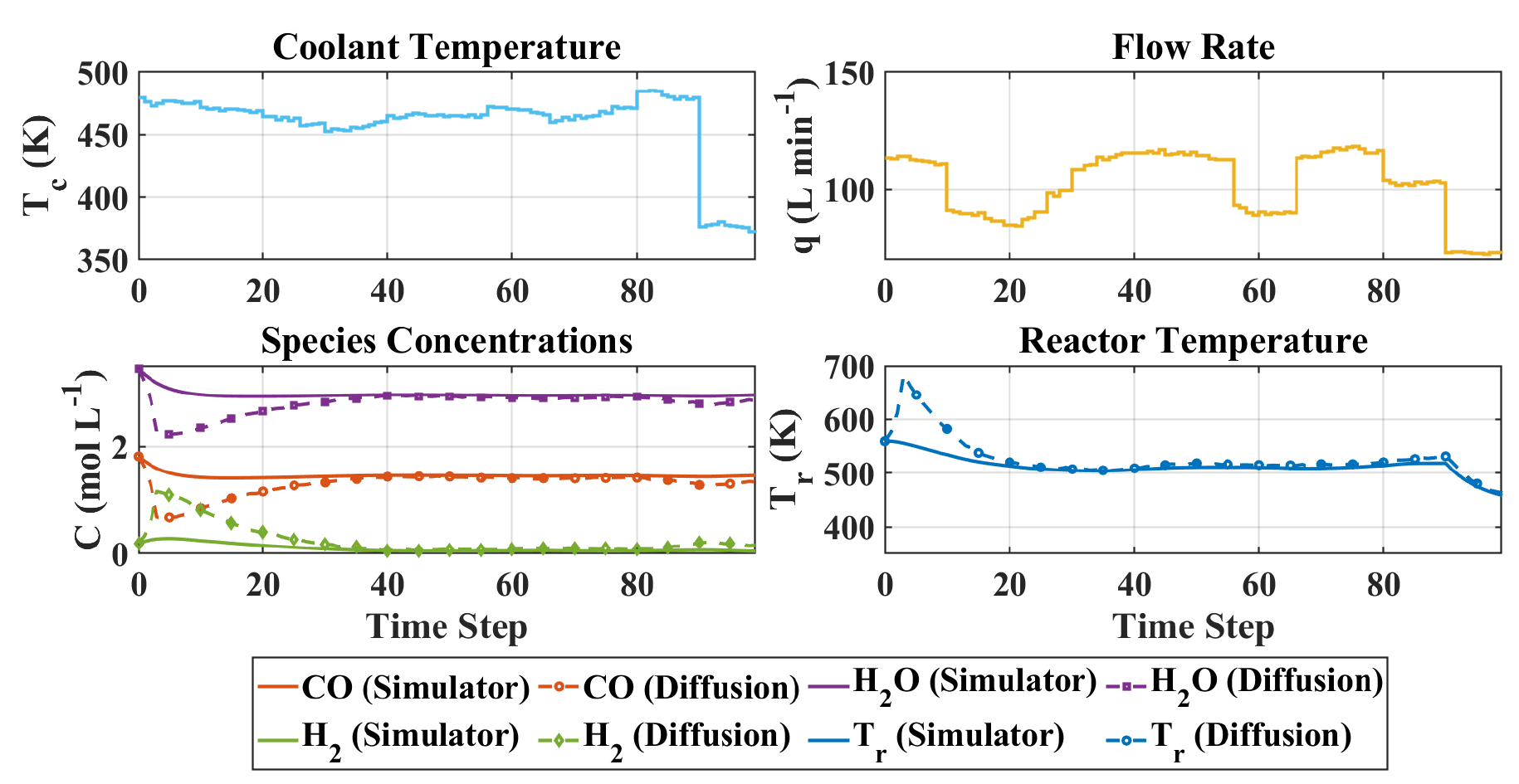}
	}
	\caption{Comparison between synthetic and ground-truth reaction profiles (extrapolation) for heat transfer efficiency $f=0.55$ and activation energy $f_{E_a}=1.55$ \textit{without} the physics-guided loss.}
	\label{fig:0.55_1.55_without_physics}
\end{figure}

\begin{figure}[thb]
	\centering
	\makebox[\textwidth][c]{%
		\includegraphics[width=1\textwidth]{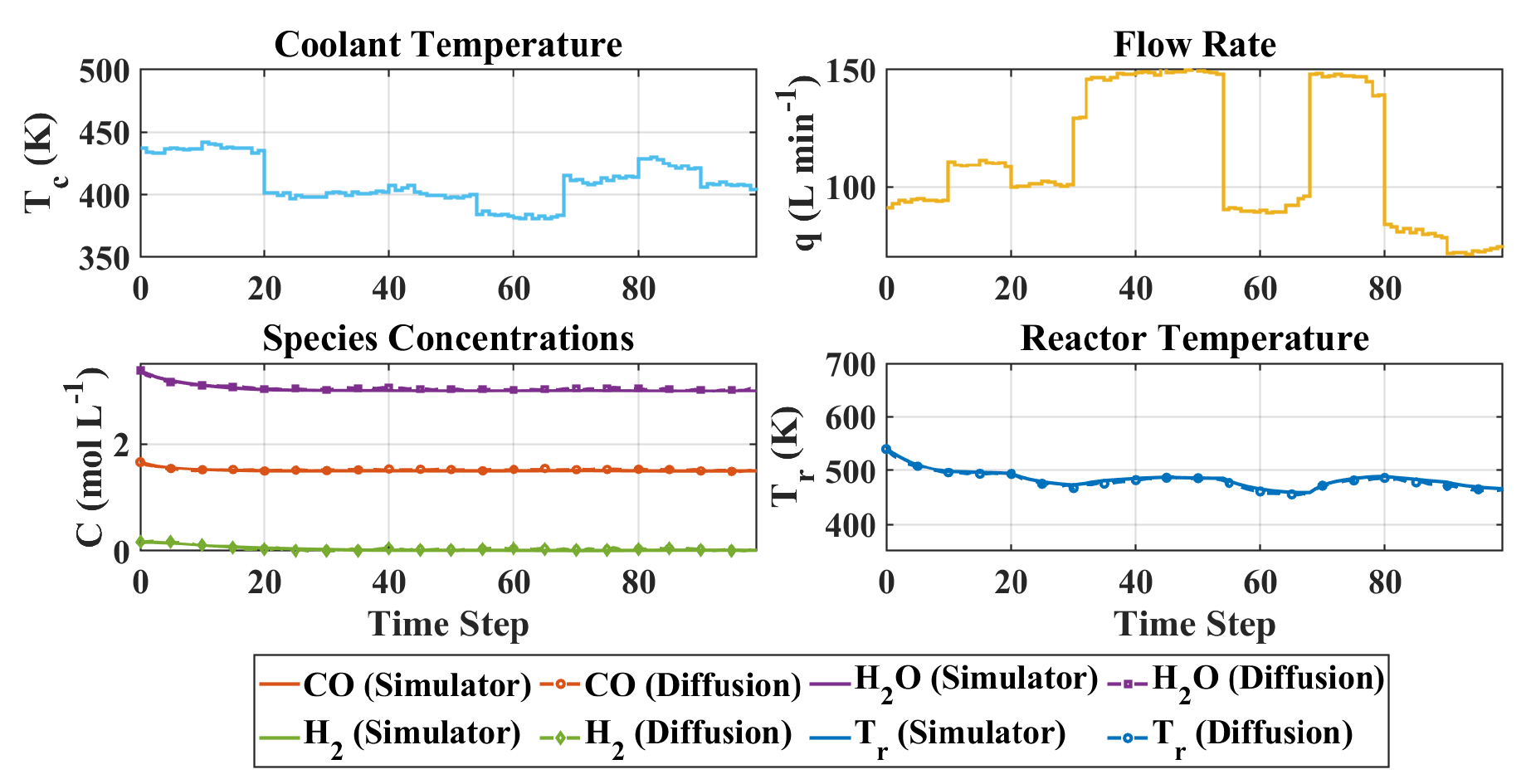}
	}
	\caption{Comparison between synthetic and ground-truth reaction profiles (extrapolation) for heat transfer efficiency $f=0.55$ and activation energy $f_{E_a}=1.55$ \textit{with} the proposed Pg-CDDPM.}
	\label{fig:0.55_1.55_with_physics}
\end{figure}

\begin{table}[h]
	\centering
	\small
	\caption{RMSE comparison for the diffusion model without and with physics-guided loss for the interpolation problem at $f=0.72$ and $f_{E_a}=1.30$.}
	\label{tab:interpolation_ablation_rmse}
	\begin{tabular}{lllll}
		\toprule
		\textbf{Model} &
		\textbf{Temperature (K)} &
		\textbf{CO} &
		\textbf{H$_2$O} &
		\textbf{H$_2$} \\
		\midrule
		Without physics-guided loss & $5.0$ & 0.0368 & 0.0377 & 0.0348 \\
		With physics-guided loss    & 4.0        & 0.0283 & 0.0264 & 0.0267 \\
		\bottomrule
	\end{tabular}
\end{table}
The effect of physics guidance is more pronounced in the extrapolation region. At $f=0.55$ and $f_{E_a}=1.55$, the model without the physics-guided loss yields a temperature RMSE of 26.6 K and concentration RMSEs of 0.2643, 0.2763, and 0.2573~mol L$^{-1}$ for CO, H$_2$O, and H$_2$, respectively, as shown in Fig.~\ref{fig:0.55_1.55_without_physics}. After incorporating the physics-guided loss, these values decrease substantially to 3.67 K and 0.0262, 0.0297, and 0.0227~mol L$^{-1}$, respectively, as shown in Fig.~\ref{fig:0.55_1.55_with_physics}. Table~\ref{tab:extrapolation_ablation_rmse} summarizes the RMSE comparison for this extrapolation case. The substantial improvement shows that the physics-guided loss helps the model generate more accurate and physically consistent trajectories under unseen degradation conditions.

\begin{table}[thb]
	\centering
	\small
	\caption{RMSE comparison without and with the physics-guided loss in the extrapolation region at $f=0.55$ and $f_{E_a}=1.55$.}
	\label{tab:extrapolation_ablation_rmse}
	\begin{tabular}{lllll}
		\toprule
		\textbf{Model} & 
		\textbf{Temperature (K)} & 
		\textbf{CO} & 
		\textbf{H$_2$O} & 
		\textbf{H$_2$} \\
		\midrule
		Without physics-guided loss & 26.6 & 0.2643 & 0.2763 & 0.2573 \\
		With physics-guided loss    & 3.67  & 0.0262 & 0.0297 & 0.0227 \\
		\bottomrule
	\end{tabular}
\end{table}

The conventional CDDPM learns only the distribution of the original training trajectories and therefore struggles to generate accurate trajectories in the extrapolation region, where no training data are available. In contrast, the proposed physics-guided loss incorporates the reaction model Eqs.~(10)–(16) to learn from both the data and governing equations. Thus it can generate more accurate trajectories outside the training region. This improvement is further demonstrated through an analysis of the RMSE distributions in both interpolation and extrapolation regions.

\subsubsection{RMSE analysis}
\begin{figure}[!ht]
	\centering
	\includegraphics[width=0.9\textwidth]{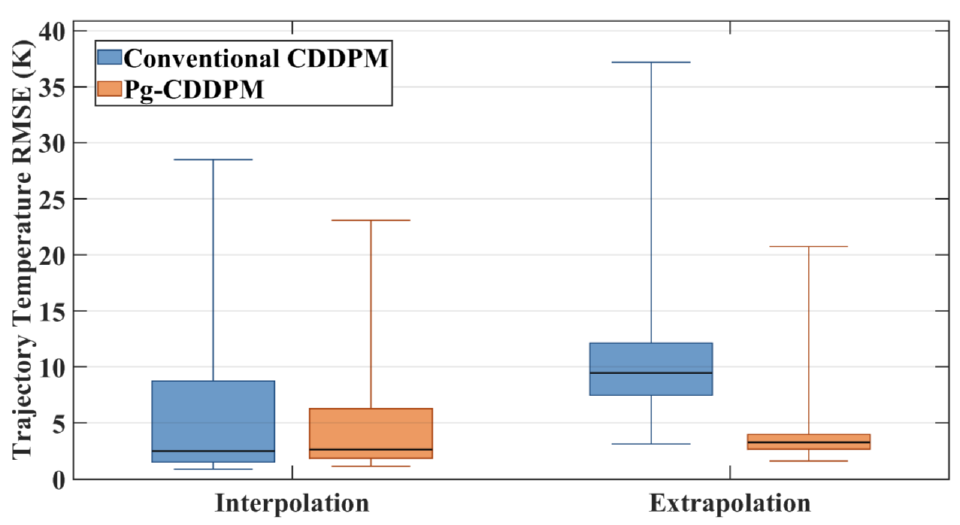}
	\caption{Comparison of the trajectory temperature RMSE distributions for the conventional C-DDPM and the proposed Pg-CDDPM under interpolation and extrapolation operating conditions.}
	\label{fig:temp_rmse}
\end{figure}

For analyzing the RMSE distributions of synthetic profiles, 500 trajectories were generated in both the interpolation and extrapolation regions using conventional data-driven CDDPM and the proposed Pg-CDDPM. Fig.~\ref{fig:temp_rmse} shows the trajectory temperature RMSE distributions under interpolation and extrapolation operating conditions. In the interpolation region, the proposed Pg-CDDPM shows a narrower error distribution than the conventional CDDPM. The improvement becomes more visible in the extrapolation region, where both the median and spread of error distributions are significantly lower for the proposed Pg-CDDPM than for the conventional CDDPM.

\begin{figure}[!ht]
	\centering
	
	\begin{minipage}[t]{0.9\textwidth}
		\centering
		\includegraphics[
		width=\linewidth
		]{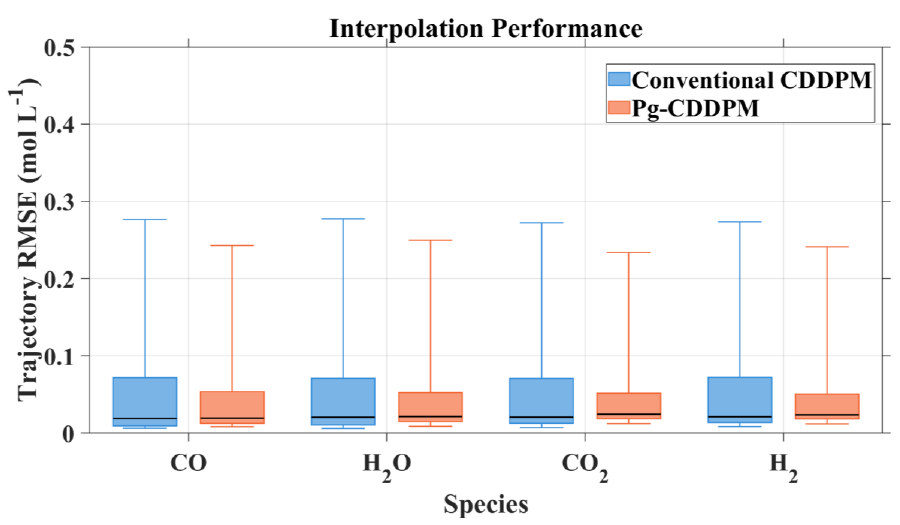}
		
		\textbf{(a)} Interpolation conditions
	\end{minipage}
	\hfill
	\begin{minipage}[t]{0.9\textwidth}
		\centering
		\includegraphics[
		width=\linewidth
		]{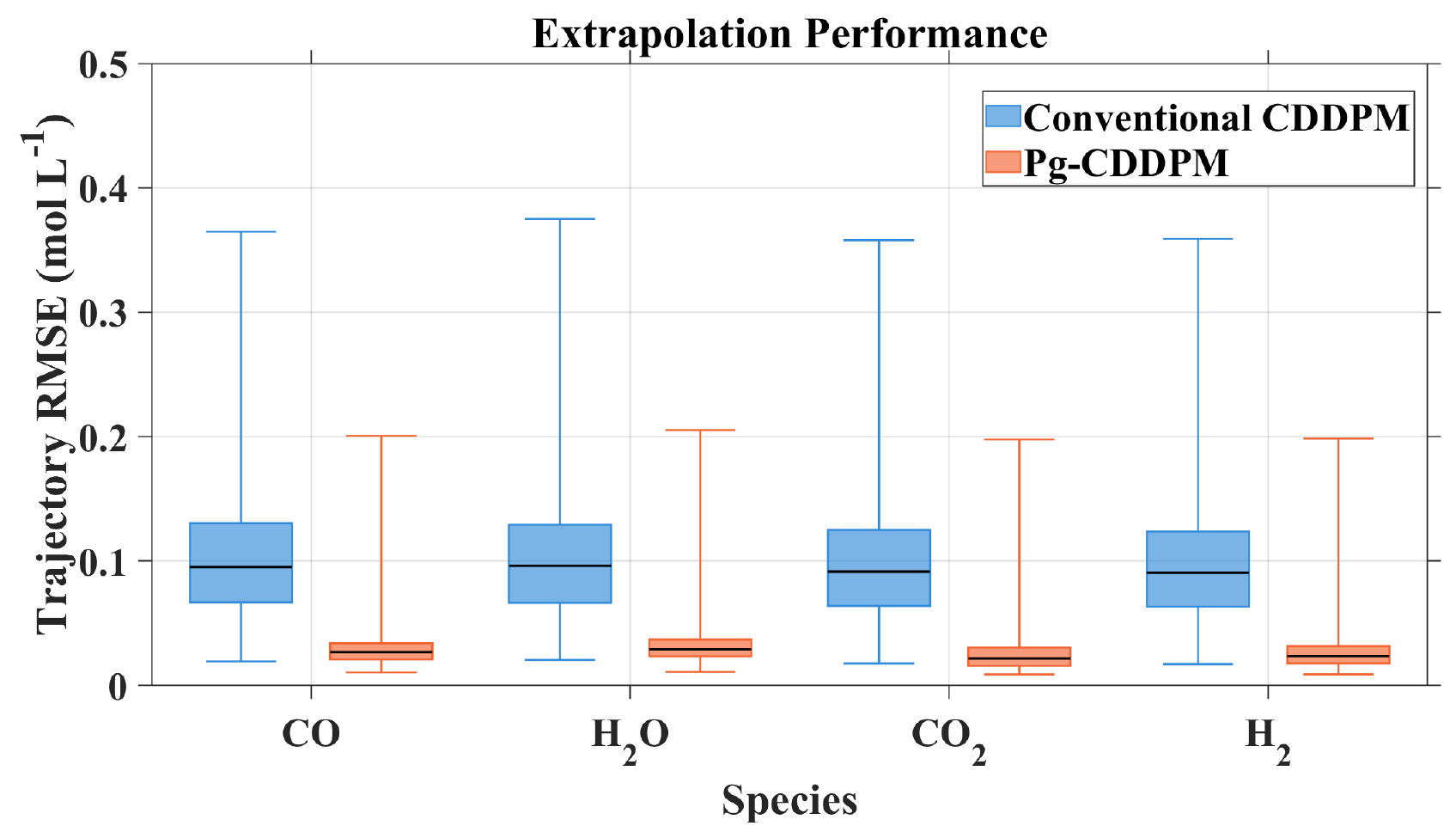}
		
		\textbf{(b)} Extrapolation conditions
	\end{minipage}
	
	\caption{Comparison of the trajectory RMSE distributions for the species concentrations under (a) interpolation and (b) extrapolation operating conditions.}
	\label{fig:conc_rmse}
\end{figure}

Figs.~\ref{fig:conc_rmse} (a) and (b) show the trajectory RMSE distributions for species concentrations in the interpolation and extrapolation regions, respectively. A similar trend is also observed in the RMSE distributions of the species concentrations. The spread of the error distribution is smaller for our Pg-CDDPM than for the conventional CDDPM, and the improvement is more visible in the extrapolation region. This demonstrates the main motivation behind using the Pg-CDDPM to generate rare-event trajectories. The overall ablation study and RMSE analysis demonstrate the contribution of the physics-guided loss to the trajectory-generation performance of the proposed model. Without including physics, the model struggles to generate accurate reactor trajectories in the extrapolation region.

\subsection{Trajectory Clustering, Augmentation, and Classifier Performance}

\subsubsection{Trajectory clustering based on the hazard score index}
As mentioned before, each of the generated trajectories is classified into one of four different operating classes based on its corresponding hazard score index.
\begin{figure}[!ht]
	\centering
	\includegraphics[width=0.9\textwidth]{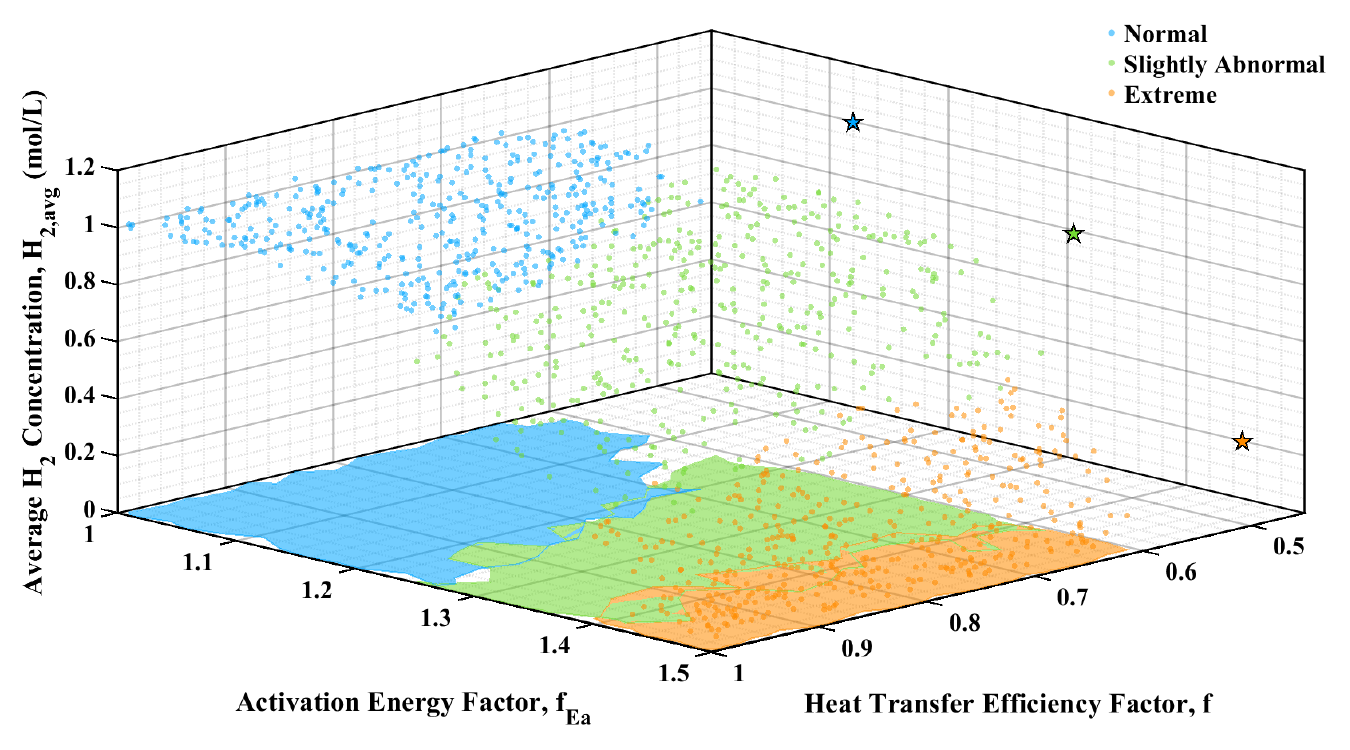}
	\caption{Clustering of the simulator-generated trajectories based on the hazard score with respect to $f$, $f_{E_a}$, and $H_{2,\mathrm{avg}}$ .}
	\label{fig:clustering_h2}
\end{figure}
Fig.~\ref{fig:clustering_h2} shows the distribution of the simulator-generated trajectories with respect to $f$, $f_{Ea}$, and average H$_2$ concentration $H_{2,\mathrm{avg}}$. As the activation energy $f$ increases (catalyst degradation) and the heat transfer efficiency $f_{Ea}$ decreases (fouling), the trajectories generally shift toward more severe operating regions, with  $H_{2,\mathrm{avg}}$ decreasing substantially. No disaster-class trajectories are observed within the original training range of $f=0.6$--$1.0$ and $f_{E_a}=1.0$--$1.5$. This represents a real-life challenge associated with the lack of rare-event data under disaster operating conditions and motivates the use of the proposed Pg-CDDPM for rare-event data augmentation and diagnosis.

\subsubsection{Trajectory augmentation}
\begin{figure}[!ht]
	\centering
	\includegraphics[width=0.95\textwidth]{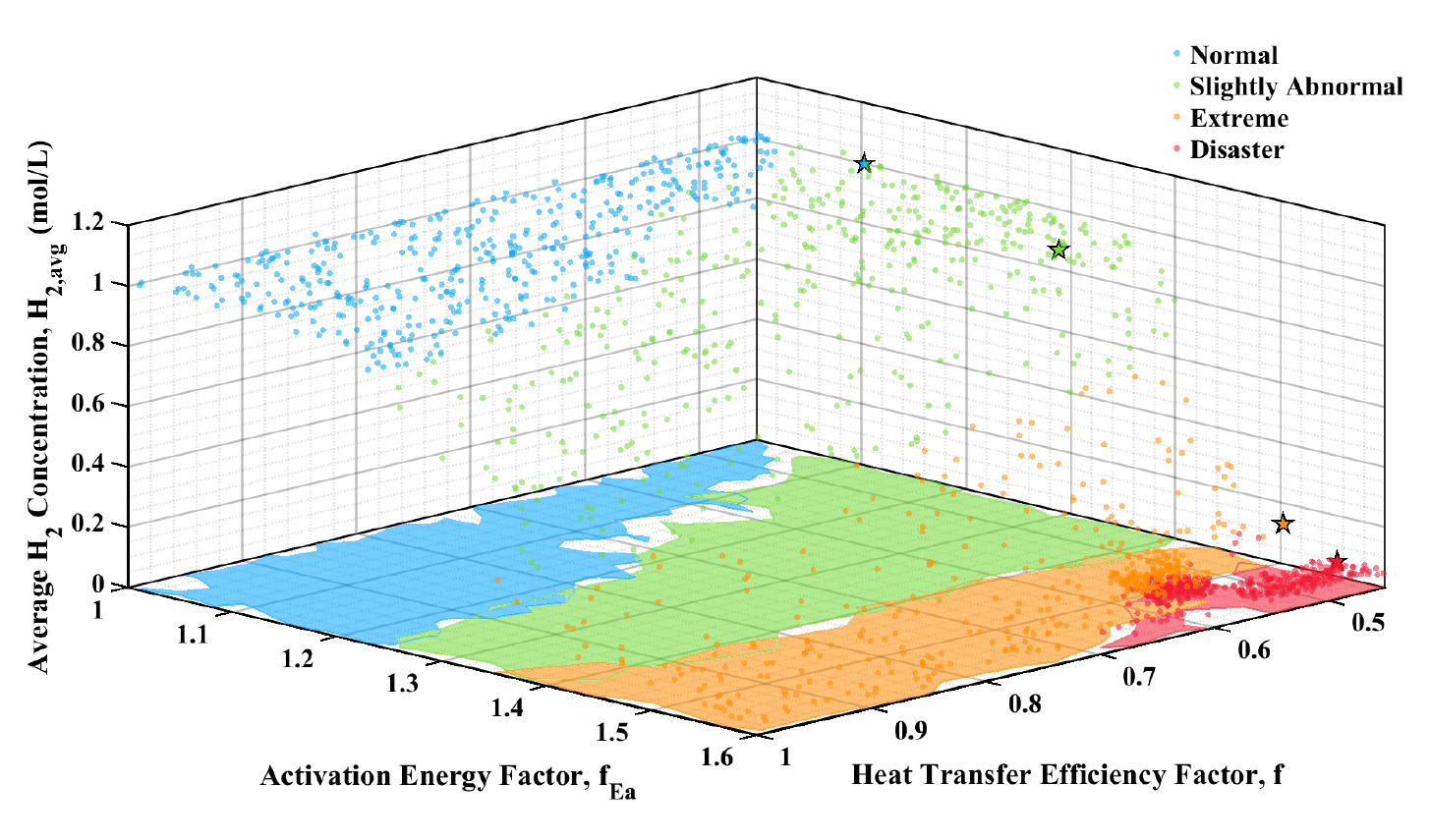}
	\caption{Distribution of the augmented trajectories based on the hazard score with respect to the $f$, $f_{E_a}$, and  $H_{2,\mathrm{avg}}$.}
	\label{fig:augmented_h2}
\end{figure}

From Fig.~\ref{fig:clustering_h2}, we can observe that the original simulator-generated trajectories cover a range of $f=0.6$--$1.0$ and $f_{E_a}=1.0$--$1.5$, and no disaster-class trajectories are observed within the training range.
To increase the amount of rare-event data, the Pg-CDDPM was used to generate additional trajectories beyond the training range. After data augmentation, the operating range was extended to $f=0.45$--$1.0$ and $f_{E_a}=1.0$--$1.6$, as in Fig.~\ref{fig:augmented_h2}. The augmented dataset contains additional trajectories in the severe operating regions, including the disaster class, providing a more balanced distribution of trajectories for training the classifier.

\subsubsection{Operating region diagnosis}
The performance of the operating region classifier, particularly structured via a GRU network, was evaluated under three training cases: the original simulator-generated dataset (Case 1), the dataset augmented using the proposed Pg-CDDPM (Case 2), and the dataset augmented using the conventional CDDPM (Case 3). The classifier uses seven process variables we considered as inputs, and outputs four operating classes: normal, slightly abnormal, extreme, and disaster. As in Table~\ref{tab:training_class_distribution}, Case 1 contains 415 trajectories for each of the normal, slightly abnormal, and extreme classes, with no disaster-class trajectories. For Cases 2 and 3, data augmentation was performed to obtain 415 trajectories for each of the four operating classes. Consequently, a total of 1,245, 1,660, and 1,660 trajectories were used for training in Cases 1, 2, and 3, respectively. A separate test dataset containing 700 trajectories was used to evaluate the classification performance. The test dataset consists of 146 normal, 185 slightly abnormal, 179 extreme, and 190 disaster trajectories. The same test dataset was used for all three cases to ensure a consistent comparison of the classification performance.


\begin{table}[!ht]
	\centering
	\small

	\caption{Class distribution of the training datasets for the three cases.}
	\label{tab:training_class_distribution}
	\begin{tabular}{llllll}
		\hline
		\textbf{Training dataset} & \textbf{Normal} & \textbf{Slight} &
		\textbf{Extreme} & \textbf{Disaster} & \textbf{Total} \\
		\hline
		Case 1: Simulation 
		& 415 & 415 & 415 & 0 & 1245 \\
		
		Case 2: Simulation + Pg-CDDPM 
		& 415 & 415 & 415 & 415 & 1660 \\
		
		Case 3: Simulation + Conventional CDDPM 
		& 415 & 415 & 415 & 415 & 1660 \\
		\hline
	\end{tabular}
\end{table}

\begin{figure}[!ht]
	\centering
	\includegraphics[width=1\textwidth]{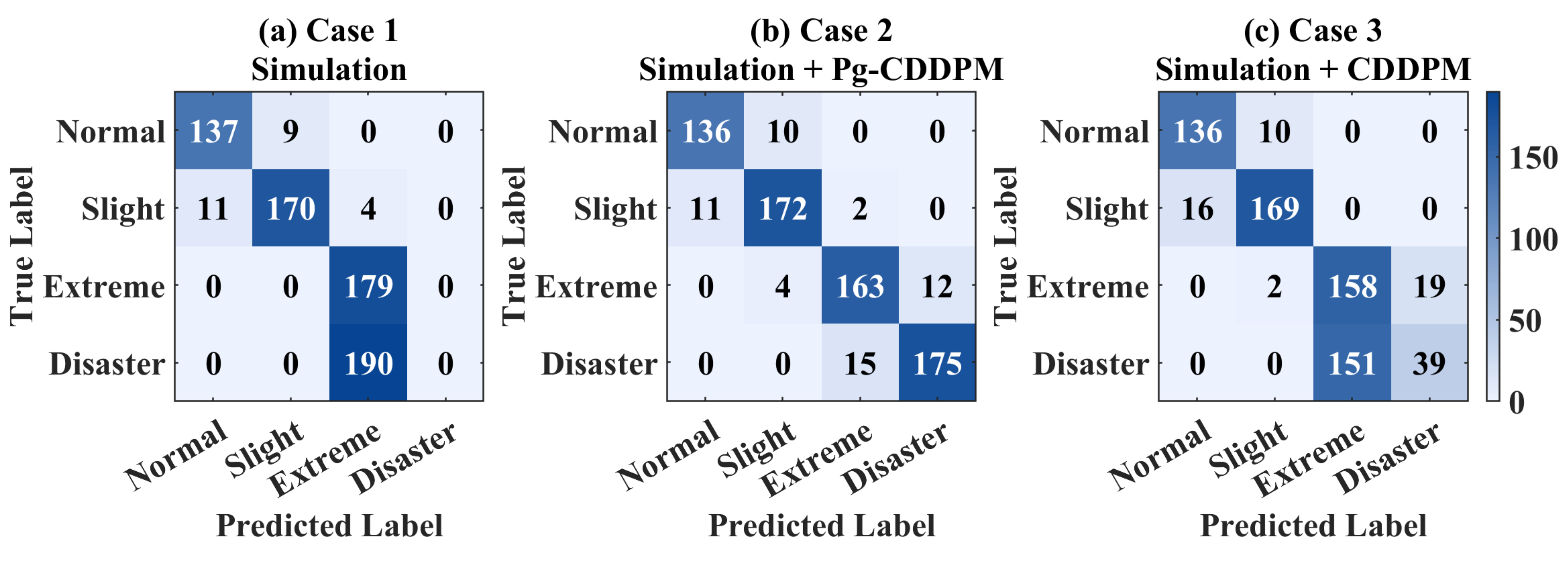}
	\caption{Confusion matrices of the GRU classifier for the three training cases.}
	\label{fig:confusion_matrix}
\end{figure}

\begin{figure}[!ht]
	\centering
	\includegraphics[width=1\textwidth]{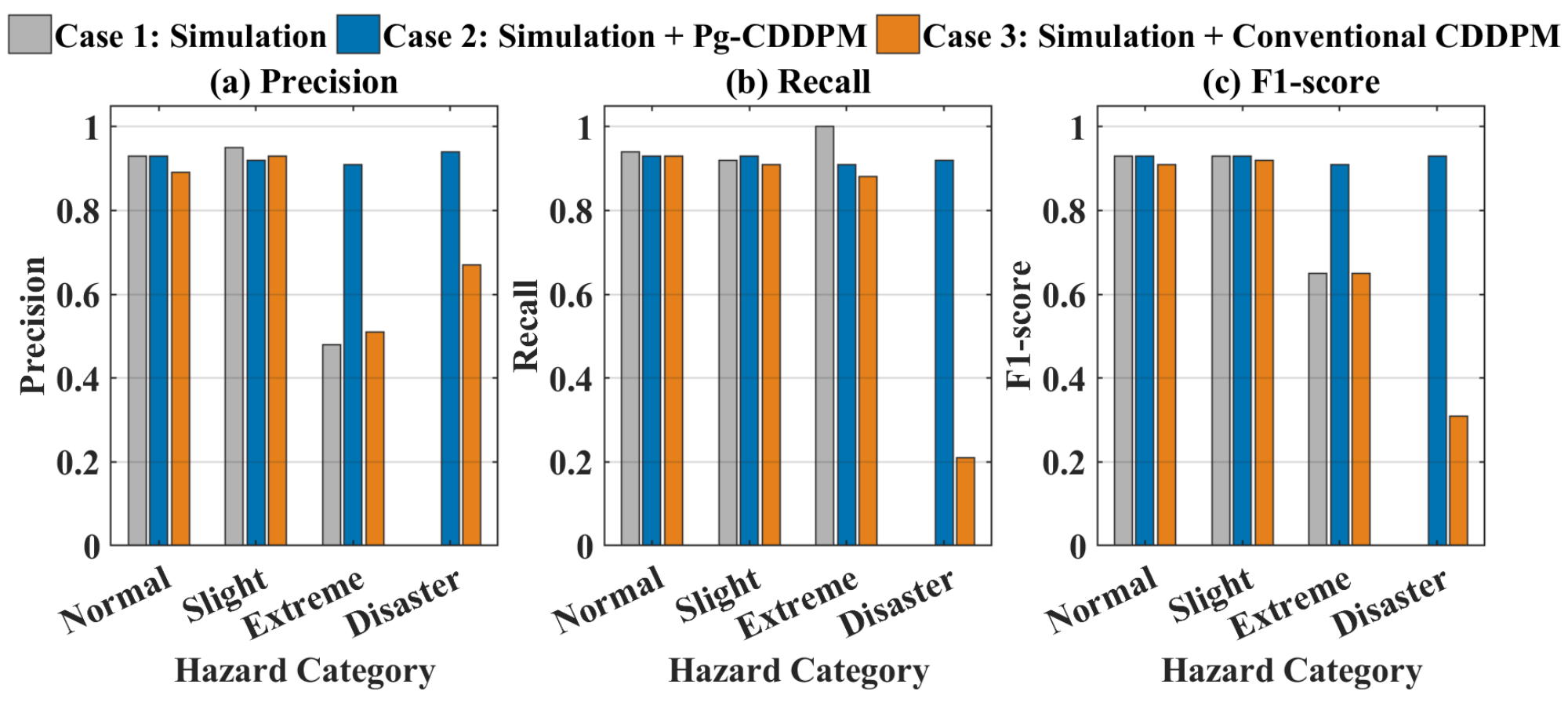}
	\caption{Comparison of the class-wise precision, recall, and F1-score obtained using the three training datasets.}
	\label{fig:performance_metrics}
\end{figure}

The diagnosis performance for three cases is compared using the confusion matrices shown in Fig.~\ref{fig:confusion_matrix}. For Case 1, the classifier performs well for the normal, slightly abnormal, and extreme classes. However, all 190 disaster trajectories are misclassified as extreme because the classifier was not exposed to disaster-class data during training. For Case 2, after augmenting the training dataset with Pg-CDDPM, the classification of the disaster class improves significantly. Out of 190 disaster trajectories, 175 are correctly classified, while only 15 are misclassified as extreme. For the extreme class, 163 out of 179 trajectories are correctly classified, with 12 misclassified as disaster and 4 grouped as slightly abnormal. For Case 3, data augmentation with CDDPM also improves the identification of the disaster class compared with Case 1. However, 151 trajectories are misclassified as the extreme class. Overall, the Pg-CDDPM augmented dataset provides better classification of the extreme and disaster conditions compared with the conventional CDDPM augmented dataset. Other classification metrics, including precision, recall, and F1-score, are presented in Fig.~\ref{fig:performance_metrics}, with their corresponding values listed in Table~\ref{tab:classification_metrics}.

\begin{table}[!ht]
	\centering
	\small
	\caption{Comparison of the classification performance for the three training cases. Case 1: Simulation; Case 2: Simulation + Pg-CDDPM; Case 3: Simulation + C-DDPM.}
	\label{tab:classification_metrics}
	
	\begin{tabular}{llllll}
		\hline
		\textbf{Metric} & \textbf{Case} & \textbf{Normal} &
		\textbf{Slightly Abnormal} & \textbf{Extreme} & \textbf{Disaster} \\
		\hline
		
		& Case 1 & 0.93 & 0.95 & 0.48 & 0.00 \\
		Precision & Case 2 & 0.93 & 0.92 & 0.91 & 0.94 \\
		& Case 3 & 0.89 & 0.93 & 0.51 & 0.67 \\
		\hline
		
		& Case 1 & 0.94 & 0.92 & 1.00 & 0.00 \\
		Recall & Case 2 & 0.93 & 0.93 & 0.91 & 0.92 \\
		& Case 3 & 0.93 & 0.91 & 0.88 & 0.21 \\
		\hline
		
		& Case 1 & 0.93 & 0.93 & 0.65 & 0.00 \\
		F1-score & Case 2 & 0.93 & 0.93 & 0.91 & 0.93 \\
		& Case 3 & 0.91 & 0.92 & 0.65 & 0.31 \\
		\hline
	\end{tabular}
\end{table}

\section{Conclusion}
In this work, we proposed a Pg-CDDPM framework for rare-event reaction profile generation and fault diagnosis for the WGS reaction. The proposed method incorporates the reactor governing laws into the diffusion training process through a physics-guided loss. Furthermore, the model was conditioned on the heat transfer efficiency factor and activation energy factor. These features are critical in determining the operating region and status of the reaction. The presence of governing laws enables reliable extension of the diffusion model to generate out-of-domain profiles for extrapolation to risky conditions where data is highly scarce or even completely absent. Numerical studies show that the generated synthetic profiles from physics-guided diffusion model outperforms those from traditional diffusion model in both interpolation and extrapolation regimes. Further, the generated rare-event trajectories were used to augment the training dataset for the extreme and disaster operating classes. With data augmentation, the operating region classifier can effectively diagnose the unseen rare events.  

\section{Acknowledgment}
The authors acknowledge the support from Texas Tech University. Md Abrar Rafid Siddique and Bibek Aryal acknowledge the Distinguished Graduate Student Assistantships (DGSA) from Texas Tech University.  



\bibliographystyle{unsrt}
\bibliography{References}

\end{document}